\documentclass[twoside]{article}

\usepackage[accepted]{aistats2024}
\usepackage[round]{natbib}

\usepackage[group-separator={,}]{siunitx} 

\usepackage[hidelinks]{hyperref} 
\usepackage{url}    
\usepackage[utf8]{inputenc} 
\usepackage[T1]{fontenc}   
\usepackage{nicefrac}
\usepackage{amsfonts}
\usepackage{mathtools}
\usepackage{booktabs} 
\usepackage{xcolor} 
\usepackage{tikz}
\usepackage{listings}
\usepackage{microtype}
\usepackage{graphicx}
\usepackage{comment}
\usepackage{multicol}
\usepackage{multirow}
\usepackage{xspace}
\usepackage[bottom]{footmisc}
\usepackage{bbm}
\usepackage{wrapfig}
\usepackage{threeparttable}
\usepackage{enumitem}
\usepackage{lipsum}
\usepackage{caption}
\usepackage{subcaption}
\usepackage[makeroom]{cancel}
\usepackage{footmisc}
\usepackage[outline]{contour}
\usepackage{anyfontsize}

\usepackage{adjustbox}
\usepackage[group-separator={,}]{siunitx} 

\definecolor{lightgrey}{rgb}{0.43,0.43,0.43}
\hypersetup{
    colorlinks,
    anchorcolor={blue!50!black},
    linkcolor={blue!50!black},
    citecolor={blue!50!black},
    urlcolor={blue!50!black}
}

\newcommand{\vsection}[1]{\vspace{-0pt}\section{#1}\vspace{-0pt}}
\newcommand{\vsubsection}[1]{\vspace{-0pt}\subsection{#1}\vspace{-0pt}}
\newcommand{\vparagraph}[1]{\vspace{-0pt}\paragraph{#1}}

\newcommand{\sparagraph}[1]{\vspace{-5pt}\paragraph{#1}}

\newcommand{\defaultttfamily}{\fontfamily{lmtt}\selectfont}
\newcommand{\mytexttt}[1]{{\defaultttfamily #1}}
\usepackage{beramono} 

\usepackage{algorithm}
\usepackage[noend]{algorithmic}

\usepackage{array}
\usepackage{eqparbox}

\usepackage{etoolbox}  
\makeatletter
\patchcmd{\algorithmic}{\addtolength{\ALC@tlm}{\leftmargin} }{\addtolength{\ALC@tlm}{\leftmargin}}{}{}
\makeatother

\allowdisplaybreaks 

\newcommand\blfootnote[1]{%
  \begingroup
  \renewcommand\thefootnote{}\footnote{#1}%
  \addtocounter{footnote}{-1}%
  \endgroup
}

\renewcommand\footnoterule{%
 \kern 5pt 
 \hrule width 0.4\columnwidth
 \kern 3.0pt 
}

\usepackage{Definitions}

\newcommand{\ito}{It\^{o}\xspace}

\newcommand\mydots{\hbox to 1em{.\hss.\hss.\hss}}

\newcommand\matern{Mat\'{e}rn\xspace}

\newcommand{\dt}{{\mathrm{d}t}}
\newcommand{\dWt}{{\mathrm{d}\WW_t}}
\newcommand{\dxt}{{\mathrm{d}\xb_t}}

\newcommand{\dom}{{\mathrm{dom}(\Acal)}}
\newcommand{\kds}{{\mathrm{KDS}(\Lcal,\mu;\Fcal)}}

\newcommand{\grouplasso}{R}

\newcommand\erdosrenyi{\text{Erd{\H{o}}s-R{\'e}nyi}\xspace}

\begin{document}

%

%

\twocolumn[

\aistatstitle{Causal Modeling with Stationary Diffusions}

\aistatsauthor{Lars Lorch \And Andreas Krause${}^*$ \And Bernhard Sch{\"o}lkopf${}^*$}

\aistatsaddress{ 
Department of Computer Science\\ETH Zürich, Switzerland \And  
Department of Computer Science\\ETH Zürich, Switzerland \And  
MPI for Intelligent Systems\\T{\"ubingen}, Germany} ]

\begin{abstract}
\vspace{-3pt}
We develop a novel approach towards causal inference. Rather than structural equations over a causal graph, we learn stochastic differential equations (SDEs) whose stationary densities model a system's behavior under interventions. These stationary diffusion models do not require the formalism of causal graphs, let alone the common assumption of acyclicity. We show that in several cases, they generalize to unseen interventions on their variables, often better than classical approaches. Our inference method is based on a new theoretical result that expresses a stationarity condition on the diffusion's generator in a reproducing kernel Hilbert space. The resulting {\em kernel deviation from stationarity (KDS)} is an objective function of independent interest.${}^\text{\hyperref[ftn:code]{1}}$
\vspace{-5pt}
\end{abstract}

\vsection{Introduction}
\vspace{-2pt}

\setcounter{footnote}{1} 

Decision-making, e.g., in the life and social sciences, requires predicting the outcomes of {\em interventions} in a system.
Causal models characterize interventions as changes to the data-generating process, 
which enables us to reason about their downstream effects.
By utilizing observed interventions, we can learn causal models that may generalize to predicting the effects of unseen interventions at test-time.

Classically, causal inference models a system $\xb \in \RR^d$ with a structural causal model (SCM) \citep{pearl2009causality}
\begin{align}\label{eq:scm}
    \xb = f(\xb, \epsilonb)  \, ,
\end{align}
where $\epsilon_j \in \RR$ are exogenous noise variables, and often under an additive noise assumption as $x_j = f_j(\xb) + \epsilon_j$.
Interventions can be realized as modifications of the functions
$\smash{f_j}$ or the noise $\epsilon_j$, and the SCM enables us to estimate the entailed distribution shifts in~$\xb$.
However, since $\smash{x_j}$ depends recursively on~$\xb$, SCMs are generally limited to modeling {\em acyclic} causal effects.

\begin{figure}
    \vspace*{0pt}
    \centering
    \hspace*{5pt}\includegraphics[width=0.97\linewidth]{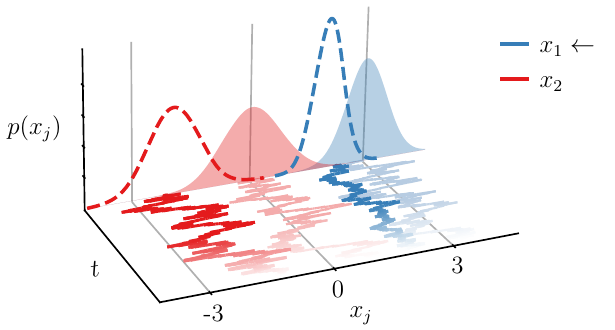}
    \vspace*{-14pt}
    \caption{
    {\bf Stationary SDEs as causal models.}
    The bottom axes show sample paths of a stationary diffusion 
    in $\smash{\RR^2}$
    before (pale) and after (dark) an intervention on the SDE governing $x_1$.
    The marginals $p(x_j)$  visualize 
    the distribution shift in $p(x_1, x_2)$.
    \vspace{-4pt}
    }
    \label{fig:illustration-intro}
\end{figure}

In this work, we propose to model a system's causal dependencies and their entailed probability distributions with stochastic differential equations (SDEs) and their entailed {\em stationary} densities. 
Specifically, we replace the SCM modeling $\xb$ by its continuous-time analogue 
\blfootnote{${}^*$Equal supervision}
\blfootnote{${}^1$%
Code: 
\label{ftn:code}\href{https://github.com/larslorch/stadion}{\fontsize{9.0}{10}\mytexttt{https://github.com/larslorch/stadion}}}
\begin{align*}
    \dxt = f(\xb_t) \dt + \sigma(\xb_t) \dWt 
\end{align*}
%
with the functions $\smash{f,\sigma}$ and the Wiener process $\{\WW_t\}$ defined later.
Stationary SDEs induce a time-invariant stationary density $\mu$ over $\RR^d$, while they internally unroll the causal dependencies of the variables over time $t$, akin to real-world processes.
We model the distribution of the variables $\xb$ by this density $\mu$, 
even though {\em not} observing the underlying process $\{\xb_t\}$ over time.
As in SCMs, interventions may be modeled as modifications to $\smash{f}$ and $\sigma$; the SDEs
characterize how the stationary density of $\xb$ changes by propagating the perturbations through its functional causal mechanisms
(Figure \ref{fig:illustration-intro}).
In the following, we will argue that modeling causation using stationary diffusions has several benefits:

\sparagraph{Cyclic systems}
Causal effects in SDEs unfold over time, so feedback loops among the variables $x_j$ become well-defined.
Feedback is ubiquitous, 
\eg,
in biological, environmental, and engineering systems \citep{hasty2002engineered,cox2000acceleration,aastrom2008feedback}, 
but SCMs a priori only allow cycles under strong model restrictions 
(\eg, \citealp{mooij2011causal,hyttinen2012learning}). 
Our results suggest that stationary diffusions are more accurate than SCMs at predicting the effects of interventions in cyclic systems, while equally competitive in acyclic settings.

\sparagraph{Graph-free}
Since acyclicity is not a constraint, stationary SDEs do not require the formalism of a causal graph. 
When learning SCMs, graphs serve as a tool for avoiding cycles in the causal dependencies, which classically restricts methods to discrete optimization (\eg, \citealp{chickering2003optimal}).
While recent works introduce continuous formulations 
of the acyclicity constraint (\eg, \citealp{zheng2018dags}), they still perform constrained optimization over the space of acyclic graphs.

\sparagraph{Flexible distribution and intervention models}
Unlike inference of causal graphs and SCMs, which often exploits the statistical properties of particular functions, exogenous noise, or interventions (\eg, \citealp{geiger1994learning,shimizu2006linear}),
our learning algorithm for stationary diffusions is general and thus agnostic to the system and intervention model.

\sparagraph{Gradient-based inference without sampling}
\looseness-1
We derive a novel kernelized objective that translates a stationarity condition on a diffusion's generator into reproducing kernel Hilbert spaces.
Using this objective, stationary SDEs can be inferred consistently via gradient-based optimization, without sampling rollouts of the model or backpropagating gradients through time.

\medskip 

To describe our approach,
we first devote Section \ref{sec:background} to 
technical 
background on kernels, SDEs, and their generators.
This material is essential for Section~\ref{sec:kds}, where we derive the {\em kernel deviation from stationarity (KDS)}.
The KDS, which later forms our foundation for inference, serves as a model-agnostic objective for fitting stationary SDEs to an empirical target density.
Building on these results,
Section \ref{sec:sdes-causal-models}  introduces stationary diffusions as causal models and describes how to infer them from interventional data using the KDS.
Sections \ref{sec:related-work} and~\ref{sec:results} conclude with related work and experiments.

\vsection{Background}\label{sec:background}

We write $\smash{\xb \in \RR^d}$ bold-faced for vectors, $\smash{x_j \in \RR}$ when indexing them, $\smash{\xb_t \in \RR^d}$ for $\xb$ at time $t$, and $\smash{\{\xb_t\}}$ for stochastic processes.
The space $\smash{C^p}$ contains all $\smash{p}$-times continuously differentiable functions $\smash{h: \RR^d \rightarrow \RR}$, and $\smash{C^p_c}$ denotes the compactly supported functions in $\smash{C^p}$.

\vparagraph{Kernels and reproducing kernel Hilbert spaces}
Throughout this work, let $k(\xb, \xb'):$~$\smash{\RR^d} \times \smash{\RR^d} \rightarrow \RR$ denote a positive definite kernel function that is four times differentiable.
Additionally, let $\Hcal$ be the reproducing kernel Hilbert space (RKHS) of functions $\smash{\RR^d} \rightarrow \RR$ 
associated with the kernel $k$ and equipped with the norm $\smash{\langle \cdot, \cdot \rangle_\Hcal}$. 
The RKHS $\Hcal$ satisfies that $k(\cdot, \xb) \in \Hcal$ for all $\xb \in \smash{\RR^d}$, where $k(\cdot, \xb)$ denotes the function obtained when fixing the second argument of $k$ at $\xb$.
Moreover, the RKHS $\Hcal$ also satisfies the {\em reproducing property} that $h(\xb) = \smash{\langle h, k(\cdot, \xb) \rangle_\Hcal}$ for all $\xb \in \smash{\RR^d}$ and $h \in \Hcal$.
In other words, evaluations of RKHS functions $h \in \Hcal$ are inner products in $\Hcal$ and parameterized by the ``feature map'' $k(\cdot, \xb)$. 
More background on kernels and RKHSs is given by
\citet{scholkopf2002learning}.

\vparagraph{Stochastic differential equations}
SDEs are a stochastic analogue to differential equations.
Rather than functions,
their solutions
are stochastic processes $\{\xb_t\}$, $\xb_t \in \smash{\RR^d}$ called diffusions,
which are sequences of random vectors indexed by $t$.
For our purposes, the Wiener process (or Brownian motion) 
$\smash{\{\WW_t\}}$, $\smash{\WW_t \in \RR^b}$ 
can be viewed as driving noise with independent increments
$\WW_{t+s}-\WW_t \sim \Ncal(0, s\Ib)$,
where usually $\smash{b=d}$.
General SDEs 
$\dxt = f(\xb_t) \dt + \sigma(\xb_t) \dWt$
with some $\xb_0 \sim p_0$
contain a drift ${f: \RR^d \rightarrow \RR^d}$ and a diffusion function ${\sigma: \RR^d \rightarrow \RR^{d \times b}}$.
We make the common assumption that 
$\smash{f}$ and $\smash{\sigma}$ are Lipschitz continuous, which ensures that the SDEs have a unique strong solution 
given the initial  vector~$\xb_0$
\citep[][Theorem 5.2.1]{oksendal2003stochastic}.
Formally, we consider integrals of $\{\WW_t\}$ under the \ito convention
\citep[][Chapter 3]{oksendal2003stochastic}.

The diffusion $\{\xb_t\}$ solving the SDEs is {\em stationary} if the probability density $\smash{\mu_t(\xb)}$ of $\xb_t$ at time $t$ is the same for all $t \geq 0$
\citep[][Chapter 4, Lemma 9.1]{ethier1986markov}.
For example, the Ornstein-Uhlenbeck process solving 
$\smash{\dxt = - \xb_t \dt + \sqrt{2} \dWt}$ has the stationary density $\mu(\xb) = \Ncal(\xb; \mathbf{0}, \Ib)$.
The so-called Fokker-Planck equation characterizes the time evolution of $\smash{\nicefrac{\partial}{\partial t} \, \mu_t(\xb)}$.
\citet{oksendal2003stochastic} provides a more formal exposition of SDEs and the Brownian motion.

\vparagraph{The infinitesimal generator}
The local evolution of a diffusion is described by its infinitesimal generator, which will play a central role in the derivation of our learning objective later on.
The generator $\Acal$ associated to a stochastic process $\{\xb_t\}$ is a linear {\em operator} that maps functions $h: \smash{\RR^d \rightarrow \RR}$ to functions of the same signature.
$\Acal$ can be viewed as the derivative of the semigroup of transition operators $\{ \Tcal_t: t \geq 0 \}$ given by $(\Tcal_t h)(\cdot) = \EE_{\{\xb_t\}}[h(\xb_t) \given \xb_0 = \cdot\,] $
and is defined as
\begin{align}\label{eq:generator}
   ( \Acal h ) (\xb) := 
    \lim_{t \downarrow 0} \frac{\Tcal_t h(\xb) - h(\xb) }{t} 
\end{align}
%
for all functions ${h \in \dom}$.
The {\em domain} $\dom$ of the generator contains all functions for which this limit exists for all $\smash{\xb \in \RR^d}$
\citep[][Chapter 1.1]{ethier1986markov}.
The operators $\{ \Tcal_t: t \geq 0 \}$ form a semigroup because $\Tcal_0$ is the identity and $\Tcal_s \Tcal_t = \Tcal_{s+t}$.
Intuitively, the generator tells us how $h(\xb_t)$ changes infinitesimally over time $t$ when $\xb_0 = \xb$ (in expectation and given an arbitrary $h$).
By Taylor's theorem, we can express $\Tcal_t h$ as
$\smash{\Tcal_t h(\zb) = h(\zb) + t\,\Acal h(\zb) + o(t)}$ for small~$t$.

If the stochastic process $\{\xb_t\}$ solves the system of SDEs $\dxt = f(\xb_t) \dt + \sigma(\xb_t) \dWt$, then its generator $\Acal$ can be expressed in terms of $f$ and $\sigma$ 
for a large class of functions $h$.
Specifically, for all functions $\smash{h \in C^2_c}$, we have $\Acal = \Lcal$ and ${h \in \dom}$, where $\Lcal$ is the linear differential operator~$\Lcal$ given by
\begin{align}\label{eq:differential-operator}
    \begin{split}
    (\Lcal h)(\xb) :=~& f(\xb) \cdot \nabla_\xb h (\xb) \\
    &+ \tfrac{1}{2} \, \mathrm{tr}\big(\sigma(\xb) \sigma(\xb)^\top \nabla_\xb \nabla_\xb h (\xb)\big) \, 
    \end{split}
\end{align}
(\citealp[][Theorem 7.3.3]{oksendal2003stochastic}).
For diagonal $\sigma$, the trace in \eqref{eq:differential-operator} reduces to a weighted sum of the unmixed second-order partial derivatives of $h$, and for $\sigma = \Ib$, to the Laplacian $\Delta_\xb h(\xb)$.

\vsection{The Kernel Deviation from Stationarity}\label{sec:kds}
\looseness-1
Suppose we are given a target density $\mu$ over $\smash{\xb \in \RR^d}$.
How can we learn the functions $\smash{f}$ and $\sigma$ of a general system of SDEs $\dxt = f(\xb_t) \dt + \sigma(\xb_t) \dWt$ such that the diffusion solving the SDEs has the stationary density~$\mu$?
In this first part, we will study this general inference question without yet considering causality and interventions in SDEs.
Our starting point 
is a well-known link between the generator of a stochastic process and its stationary density.
For a stochastic process $\{\xb_t\}$,
the density $\mu$ is the stationary density of $\{\xb_t\}$ if and only if the generator $\Acal$ 
associated to $\{\xb_t\}$ satisfies
\begin{align}\label{eq:generator-stationary-result}
    \EE_{\xb \sim \mu} \big[ \Acal h (\xb) \big] = 0
\end{align}
for all functions $h$ in a {\em {core}} for the generator $\Acal$ 
\citep[][Chapter 4, Proposition 9.2]{ethier1986markov}.
Roughly speaking, a core is a dense subset of functions in the domain $\dom$ such that, if \eqref{eq:generator-stationary-result} holds for all functions $h$ in the core, 
then \eqref{eq:generator-stationary-result} also holds for $h \in \dom$ \citep[see][Section 6]{hansen1995back}.
Equation \eqref{eq:generator-stationary-result} states that every function $h$ of $\{\xb_t\}$ must have zero rate of change $\smash{\Acal h(\xb)}$ in expectation over initializations by the stationary density $\mu$.
In other words, any $\smash{h(\xb_t)}$ must be, in expectation, invariant with time~$t$ iff $\mu$ is stationary.



If we can verify that the expected infinitesimal change over a target density $\mu$ is zero for an expressive class of test functions $h$
(or conversely, learn a system of SDEs satisfying this),
we may conclude that $\mu$ is a stationary density of the solution $\smash{\{\xb_t\}}$ to the SDEs.
This insight suggests that it is sufficient to find the function $\smash{w}$ achieving the {\em largest} deviation from $\smash{\EE_{\xb \sim \mu} [ \Acal h (\xb) ]} = 0$
among all test functions $h$.
In the following, we derive a closed form for this maximum deviation over a sufficiently-rich, {\em infinite} set of functions as well as the witness function $w$ achieving this maximum.
We sketch our proofs and defer their formal arguments to Appendix~\ref{app:proofs}.

\vsubsection{Bounding the Deviation from Stationarity}\label{ssec:kds-upper-bounding}
Our key idea for bounding the functional in \eqref{eq:generator-stationary-result} is to consider functions $h$ in an RKHS $\Hcal$.
We show that this allows us to derive a closed-form expression for the supremum of  \eqref{eq:generator-stationary-result} over an expressive, infinite subset of functions in the RKHS.
In the following, let $\Hcal$ be the RKHS of a kernel $k$
as introduced in Section \ref{sec:background},
and let $\Fcal := \{ h \in \Hcal :  \lVert h \rVert_{\Hcal} \leq 1 \}$ be the unit ball of~$\Hcal$.

To begin, we first focus on the closely-related functional $\EE_{\xb \sim \mu}[\Lcal h(\xb)]$ involving the operator $\Lcal$ instead of the generator $\Acal$ (Section~\ref{sec:background}).
Recall that the operator $\Lcal$ coincides with the generator $\Acal$ of the diffusion $\smash{\{\xb_t\}}$ solving $\dxt = f(\xb_t) \dt + \sigma(\xb_t) \dWt$ when applied to the well-behaved functions $\smash{C^2_c}$.
For this functional, we can show that there exists a representer function $\smash{g_{\mu,\Lcal}}$ in the RKHS $\Hcal$, 
whose inner product with any function $\smash{h \in \Hcal}$ allows evaluating the functional:
\begin{lemma}\label{lemma:embedding}
Let $\mu$ be a probability density over $\RR^d$ and assume 
that the functions 
$f$, $\sigma$, and the partial%
$\,$\footnote{Like \citet{steinwart2008support}, we use $\nicefrac{\partial}{\partial x_{i,i}}$ to denote the first-order partial derivative w.r.t.\ both function arguments, so
$\nicefrac{\partial}{\partial x_{i,i}} k(\xb, \xb) := 
\nicefrac{\partial}{\partial u_i} \nicefrac{\partial}{\partial v_i} k(\ub, \vb)\vert_{\ub := \xb, \vb := \xb}$.%
\label{footnote:partial}}
derivatives $\nicefrac{\partial}{\partial x_{i,i}} k(\xb, \xb)$ and 
$\smash{\nicefrac{\partial^2}{\partial x_{i,i}\partial x_{j,j}} k(\xb, \xb)}$
are square-integrable with respect to $\mu$. Then, there exists a unique function $g_{\mu,\Lcal} \in \Hcal$ satisfying
\begin{align*}
    \EE_{\xb \sim \mu}[\Lcal h(\xb)] = \langle h, g_{\mu,\Lcal} \rangle_\Hcal
\end{align*}
for any $h \in \Hcal$. 
Moreover,
$g_{\mu,\Lcal}(\cdot) = \EE_{\xb \sim \mu}[\Lcal_\xb k(\xb, \cdot\,)]$.
Here, the notation $\Lcal_\xb$ indicates that $\Lcal$ is applied to the argument $\xb$.
\end{lemma}
To prove Lemma \ref{lemma:embedding}, we show that $\EE_{\xb \sim \mu}\Lcal$ is a continuous linear functional
and then invoke Riesz' representation theorem to prove that $\smash{g_{\mu,\Lcal}}$ exists. 
The reproducing property of $\Hcal$ yields the explicit form of $\smash{g_{\mu,\Lcal}}$.
Crucially, the representation in 
Lemma~\ref{lemma:embedding}
allows us to derive a closed form for the supremum of $\smash{\EE_{\xb \sim \mu}[\Lcal h(\xb)]}$  
over the unit ball $\Fcal$, because the inner product with functions of $\Fcal$ is maximized by the unit-norm function aligned with $\smash{g_{\mu,\Lcal}}$, that is, by $w_{\mu,\Lcal} := g_{\mu,\Lcal} / \lVert g_{\mu,\Lcal} \rVert_\Hcal$.
Their inner product is then $\langle g_{\mu,\Lcal} / \lVert g_{\mu,\Lcal} \rVert_\Hcal, g_{\mu,\Lcal} \rangle_\Hcal = \lVert g_{\mu,\Lcal} \rVert_\Hcal$.
We will refer to the square of this RKHS norm as the {\em kernel deviation from stationarity} $\smash{\kds}$:

\begin{theorem}\label{theorem:sup}
Under the assumptions of Lemma \ref{lemma:embedding}, 
\begin{align*}
    \sup_{h \in \Fcal}  \EE_{\xb \sim \mu} \big[ \Lcal h (\xb) \big] = 
    \sqrt{\kds} \, ,
\end{align*}
where $\kds := 
\smash{\EE_{\xb \sim \mu} [ \Lcal_{\xb} \, \EE_{\xb' \sim \mu} [\Lcal_{\xb'} k(\xb, \xb') ] ]}.$
Under additional regularity conditions on the functions $f,\sigma,k,\text{and }\mu$, 
we may interchange the limits involved in the differentiation and integration operators and write
$\kds = \smash{
\EE_{\xb \sim \mu, \xb' \sim \mu} \big [ \Lcal_{\xb\vphantom{'}}\Lcal_{\xb'\vphantom{'}} k(\xb, \xb') \big ]}$. 
\end{theorem}
%
When thinking of $\Lcal$ as the generator $\Acal$, the witness $\smash{w_{\mu,\Lcal}}$ is the smooth RKHS function that is 
subject to the largest infinitesimal change in the diffusion when initialized in expectation over~$\mu$; 
the functions in $\Fcal$ are smooth, since their RKHS norm is limited to \num{1}.
Moreover, the KDS measures the maximal absolute deviation from \eqref{eq:generator-stationary-result} of any function 
in $\Fcal$.
While ${\EE_{\xb \sim \mu} [ \Lcal h (\xb) ]}$
can be negative, ${\sup_{h \in \Fcal}  \EE_{\xb \sim \mu} [ \Lcal h (\xb) ]}$ is always nonnegative, not only because it is equal to the RKHS norm $\lVert g_{\mu,\Lcal} \rVert_\Hcal \geq 0$.
By the linearity of the functional $\Lcal$,
we also have $\smash{\EE_{\xb \sim \mu} [ \Lcal (-h) (\xb) ]}
= \smash{- \EE_{\xb \sim \mu} [ \Lcal h (\xb) ]}$,
and since $\smash{h \in \Fcal}$ iff $\smash{-h \in \Fcal}$, the supremum over $\Fcal$ is nonnegative.

The KDS thus expresses the maximum discrepancy between $\Lcal$ and a target $\mu$ in a kernelized, closed form over $\Fcal$.
We leverage this perspective later for learning stationary SDEs from data, 
since the SDE functions $f$ and $\sigma$ enter the KDS through the operator $\Lcal$.
Before doing so, we investigate whether the KDS of an RKHS $\Hcal$ sufficiently discriminates among SDE models.

\vsubsection{Consistency}
\looseness-1
While the KDS measures a deviation from stationarity, 
it may not be consistent---%
$\kds = 0$ may not guarantee that all SDEs entailing the operator $\Lcal$ indeed induce the stationary density $\mu$.
Guaranteeing this requires that
the equality of the functional of $\Acal$ in \eqref{eq:generator-stationary-result} holds
for all functions in a core for $\Acal$.
However, the SDE-parameterized operator $\Lcal$ only coincides with the generator $\Acal$ 
of the diffusion 
for all $\smash{h \in C^2_c}$ (Section \ref{sec:background}).
Moreover, $\Fcal$ may not be dense in a core for $\Acal$ and thus fail to be sufficiently rich for testing the condition in \eqref{eq:generator-stationary-result}.

To link the KDS to $\Acal$, we need to relate a core for $\Acal$ to the functions spanned by the RKHS $\Hcal$.
In general, the relationship between these two function spaces strongly depends on the generality of the functions $f,\sigma$ defining the SDEs \citep[][Chapter 8]{ethier1986markov} and the kernel $k$ \citep{christmann2010universal,kanagawa2018gaussian}.%
\footnote{\looseness-1 Universal kernels \citep{micchelli2006universal} are of limited use here, since the denseness of universal RKHSs is usually established in supremum norm, not for the partial derivatives in $\Lcal$,
and diffusions are defined over $\smash{\RR^d}$ (noncompact).}
In the following, we show the consistency of the KDS 
for the \matern kernel $\smash{k_{\nu,\gamma}}$,
which generalizes the Gaussian kernel $\smash{k_{\gamma}(\xb, \xb') = \exp(-\lVert \xb - \xb' \rVert_2^2 /2\gamma^2)}$
(Appendix \ref{app:additional-background}).
We achieve this by showing that a core for $\Acal$ is dense in the \matern RKHS with respect to a Sobolev norm.
Building on this, we then prove that, for any $h$ in the core, there always exists a nearby RKHS element ensuring that $\smash{\EE_{\xb \sim \mu} [\Acal h(\xb)]}$ is arbitrarily small:
\begin{theorem}\label{theorem:consistency}
    Let $\smash{k_{\nu,\gamma}}$ be a \matern kernel with $\smash{\nu > 2}$ defined over $\RR^d$, and let $\Fcal$ be the unit ball of its RKHS.
    Let $\mu$ be a probability density over $\RR^d$ 
    and $f, \sigma$ be bounded functions with $\sigma\sigma^\top$ positive definite that define the SDEs $\smash{\dxt = f(\xb_t) \dt + \sigma(\xb_t) \dWt}$.
    Then, $\mu$ is a stationary density of the stochastic process $\{\xb_t\}$ solving the SDEs if and only if 
    \begin{align*}
        \kds = 0 \, .
    \end{align*}
\end{theorem}
Our result shows that the KDS is zero if and only if a system of SDEs with bounded functions $f$ and $\sigma$ induces the stationary density $\mu$.
The boundedness assumption allows us to leverage established results in stochastic analysis on cores of generators of diffusions.
However, more general results on cores may exist and imply the consistency of the KDS for other classes of kernels or SDEs.
In applications, we usually work with continuous functions $f$ and $\sigma$, which
can only be unbounded as $\smash{\lVert \xb \rVert \rightarrow \infty}$.
In practice, this occurs with arbitrarily low probability if a system is stationary,
so we may think of Theorem \ref{theorem:consistency} as informally extending 
to the case of continuous functions, even though not formally covered by the assumptions.

It is worth noting that SDEs may not have {\em unique} stationary densities in general. 
Different initial distributions of the random vector $\xb_0$ may result in different stationary densities of the solution $\{\xb_t\}$. The Lipschitz assumptions in Section \ref{sec:background} only guarantee that $\{\xb_t\}$ is the unique strong solution given $\xb_0$.
From our inference viewpoint, however, we always initialize $\xb_0 \sim \mu$ with the target $\mu$, which ensures that our SDE models have unique stationary densities.
Under additional assumptions, stationary densities can be shown to be unique for any $\xb_0$
(see \citealp{khasminskii2011stochastic}, Section 4.4).

\begin{figure*}[t]
    \centering
    \vspace*{-2pt}
    \hspace*{-6pt}\includegraphics[width=1.02\linewidth]{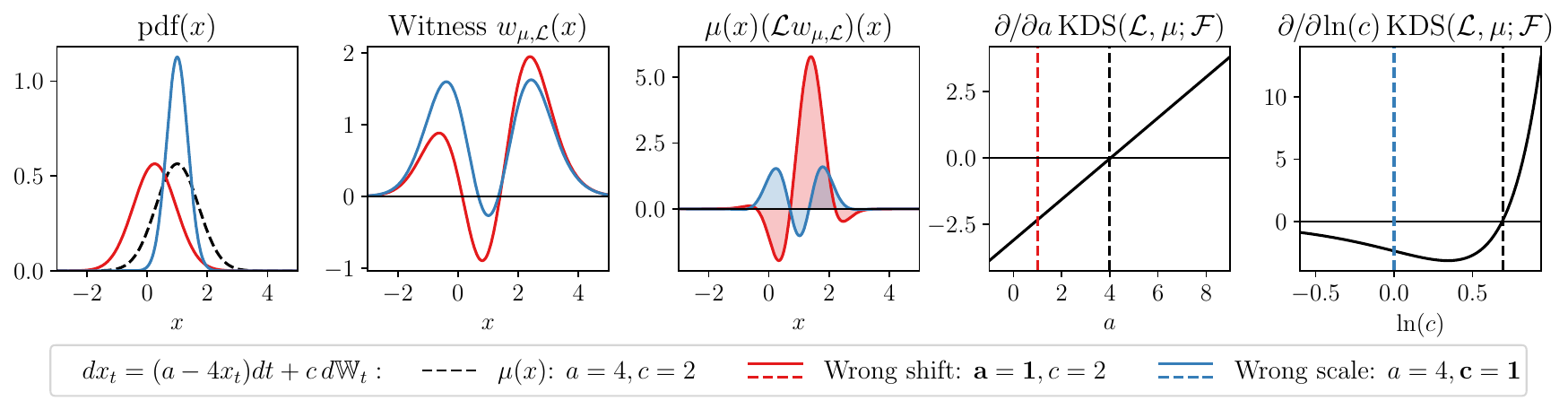}
    \vspace*{-12pt}
    \caption{%
    \looseness-1
    {\bf Components of the KDS for a stationary linear SDE}
    and a Gaussian kernel $\smash{k_\gamma}$ with $\gamma = 0.5$.
    Expectations over $\mu$ are approximated with $1000$ samples.
    {\em 1:}~%
    Densities of a target ($\mu$, black) and two alternative models. 
    {\em 2:}~%
    KDS witness functions for the misspecified models.
    {\em 3:}~%
    Witnesses after applying $\Lcal$, yielding their time derivatives in the diffusion.
    After multiplying by $\mu$, the KDS is equal to the integral of the shaded areas.
    {\em 4-5:}~%
    KDS derivatives with respect to $a$ and $c$, fixing the other parameters at those of the target model.
    The partial derivatives have zeroes at the true parameters of the model inducing $\mu$, thus
    gradient descent drives the incorrect $a$ and $c$ to their true values
    (indicated by vertical, dashed lines).
    }
    \label{fig:witness-main}
\end{figure*}

\vsubsection{The KDS as a Learning Objective}
The $\mathrm{KDS}$ provides a closed-form expression for the maximum stationarity violation of any $\smash{h \in \Fcal}$. 
Since it quantifies this violation (as an RKHS norm), the KDS serves as an objective we can minimize to fit a system of SDEs to a target density $\mu$.
Specifically, given a dataset $\smash{D = \{\xb^{(1)}, \mydots, \xb^{(N)}\}}$
of i.i.d.\ samples $\smash{\xb^{(n)} \sim \mu}$,
we can compute an unbiased empirical estimate of the $\kds$ with the U-statistic given by 
%
\begin{align}\label{eq:kds-sampled}
    \hspace*{-2pt}
    \hat{\mathrm{KDS}}(\Lcal, D; k) := 
    \tfrac{1}{N(N-1)} \sum_{m=1}^{N} \sum_{n\neq m}^{N} 
    \Lcal_{\xb}\Lcal_{\xb'} k(\xb^{(m)}, \xb^{(n)}) 
    \vspace*{-12pt}
\end{align}
In Appendix \ref{app:details-kds-linear}, we also provide an unbiased estimate of the KDS that scales {\em linearly} in $N$, which can be useful in large-scale applications.

When the SDE model $\smash{f_{\thetab}, \sigma_{\thetab}}$ is parameterized by $\thetab$, we will indicate the dependency of the operator $\Lcal$ by a superscript (here as $\Lcal^{\thetab}$).
The KDS depends on the SDE parameters $\thetab$ through the operator $\Lcal^{\thetab}$ since $\Lcal^{\thetab} h(\xb) = f_{\thetab}(\xb) \cdot \nabla_\xb h(\xb)+ \tfrac{1}{2} \mathrm{tr}(\sigma_{\thetab}(\xb) \sigma_{\thetab}(\xb)^\top \nabla_\xb \nabla_\xb h(\xb) ) $. 
Minimizing the KDS thus enables us to estimate the parameters of a stochastic dynamical system without backpropagating gradients through time.
Moreover, the function $\Lcal^{\thetab}_{\xb}\Lcal^{\thetab}_{\xb'} k(\xb, \xb')$ inside the KDS is fully differentiable with respect to $\thetab$.
Notably, the KDS is exact up to the sample approximation of the expectations over the target~$\mu$ in \eqref{eq:kds-sampled}---there are no SDE model components we need to sample from, roll out, reparameterize, or approximate.

It is instructive to consider the special case of $\smash{\sigma = \Ib}$.
The function $\Lcal^{\thetab}_{\xb}\Lcal^{\thetab}_{\xb'} k$ inside the KDS is then given by
\begin{align}\label{eq:generator-explicit-form}
\begin{split}
    \Lcal^{\thetab}_{\xb}\Lcal^{\thetab}_{\xb'} k(\xb, \xb') 
    &= f_{\thetab}(\xb) \cdot \nabla_{\xb} \nabla_{\xb'} k(\xb, \xb') \cdot f_{\thetab}(\xb')\\
    &+ \tfrac{1}{2} f_{\thetab}(\xb) \cdot \nabla_{\xb} \Delta_{\xb'} k(\xb, \xb') \\
    &+ \tfrac{1}{2} f_{\thetab}(\xb') \cdot \nabla_{\xb'} \Delta_{\xb} k(\xb, \xb')\\
    &+ \tfrac{1}{4} \Delta_{\xb}\Delta_{\xb'} k(\xb, \xb') \, , 
\end{split}
\end{align}
where $\Delta_{\xb} := \tr \nabla_\xb \nabla_\xb$ is the Laplacian.
This expression contains a matrix, two vectors, and a scalar involving $k$ that are all independent of $\thetab$.
The kernel terms may thus be reused, \eg, during gradient descent on $\thetab$.
For general $\sigma_{\thetab}$, there also exists an explicit expression, but it may be easier to leverage the operator view of $\smash{\Lcal^{\thetab}}$ to compute $\Lcal^{\thetab}_{\xb}\Lcal^{\thetab}_{\xb'} k$ and its gradients with automatic differentiation.
We provide the explicit form and pseudocode demonstrating this case
in Appendix \ref{app:details-kds-explicit}.

\vspace*{-2pt}
\vparagraph{Example}
Figure \ref{fig:witness-main} illustrates how the KDS may be used to learn the SDE parameters $\thetab$.
We consider a target model $dx_t = (a + bx_t)dt + c\dWt$ with the closed-form density $\mu(x) = \Ncal(x; -a/b, -c^2/2b)$ for $\smash{b < 0}$ and $\smash{c > 0}$ \citep{jacobsen1993brief}.
We use the KDS, approximated by samples from $\mu$, to measure the fit of two models with incorrect $a$ and $c$ controlling the mean and variance, respectively.
The partial derivatives of the KDS have zeroes at the true parameters of the model inducing $\mu$ and can thus be inferred with gradient descent.

\vsection{Stationary Diffusions as Causal Models}\label{sec:sdes-causal-models}

In this section, we describe how stationary diffusions can serve as causal models.
To facilitate this exposition, we first leave the KDS aside and focus on discussing causality in SDEs, intervention models, and related properties. 
To conclude, we then leverage the KDS as an objective for learning stationary diffusions as causal models from a collection of interventional datasets.

\vsubsection{Modeling Causal Dependencies with Stationary SDEs}
\looseness-1
Probabilistic causal models of a system $\smash{\xb \in \RR^d}$ entail more than the {\em observational} density of the variables.
A causal model contains additional information that characterizes the {\em interventional} densities of the system under interventions on its data-generating process \citep{peters2017elements}.
This information may be in the form of, say, functions $\smash{f_j}$ that explicitly model the densities of $x_j$ and remain invariant under interventions elsewhere, as in SCMs.
Which causal model of a system and which intervention model are adequate depends on the application and the level of modeling granularity \citep{hoel2013quantifying,rubenstein2017causal,scholkopf2022causality}.

\looseness-1
In this work, we propose to characterize the causal dependencies among the variables $\xb$ by modeling a stationary dynamical system $\xb_t \in \RR^d$ underlying the generative process of $\xb$. 
Specifically, we model the distribution of the variables $\xb$ with the stationary density $\mu$ of a process $\xb_t$, which evolves according to the SDEs
\begin{align}\label{eq:sde-param}
    \dxt = f_{\thetab}(\xb_t) \dt + \sigma_{\thetab}(\xb_t) \dWt \, ,
\end{align}
where $\smash{\thetab \in \RR^k}$ are parameters
and $\mu$ is the observational stationary density, \ie, $\xb_t \sim \mu$.
The core idea is that introducing an explicit time dimension enables propagating feedback cycles in the causal dependencies of the variables---even though we do {\em not} observe the system $\xb$ itself as a time series.
Through stationarity, time remains remains internal to the SDE model.
In contrast to stationary SDEs, SCMs do not allow for cycles in the causal structure except under restrictive model and invertibility assumptions (see {\em Related Work} in Section \ref{sec:related-work}).

Similar to the structural equations in SCMs, the differential equations in SDEs provide a {\em mechanistic} (or functional) model of the causal dependencies among the variables $\xb$ \citep{peters2017elements,scholkopf2022causality}.
In other words, $\smash{f_j}$ and $\smash{\sigma_j}$ model which variables in $\xb$ affect the variable $\smash{x_j}$ via an explicit functional dependency that holds independent of perturbations of the variables or the functions governing the other variables.
However, both SCMs and stationary SDEs should be thought of as phenomenological abstractions of the true physical processes underlying the observables $\smash{\xb}$ \citep[\eg,][Section 2.3.3]{peters2017elements}, with stationary SDEs characterizing the processes explicitly over time.


Following SCMs, we may graphically summarize the causal dependencies in stationary SDEs by reading off its direct functional dependencies, \eg, to gain qualitative insights or incorporate prior knowledge.
However, neither modeling nor inference with stationary SDEs requires such a graph.
Moreover, the graphs also imply different properties than for SCMs, \eg, the modeled distribution $\mu$ is not necessarily Markovian with respect to the graph \citep[cf.][Proposition 6.31]{peters2017elements}.

\vsubsection{Interventions}
Interventions in SDEs can be modeled in various ways and often in analogy to SCMs \citep{eberhardt2007interventions}.
We formalize an intervention as transforming ${f_{\thetab}}$ and ${\sigma_{\thetab}}$ into the modified mechanisms
${f_{\thetab,\phib}}$ and ${\sigma_{\thetab,\phib}}$
with additional parameters $\phib$ such that \eqref{eq:sde-param} evolves as
$\dxt = f_{\thetab,\phib}(\xb_t) \dt + \sigma_{\thetab,\phib}(\xb_t) \dWt$
and induces the stationary density $\smash{\mu_{\phib}}$.
For example, some real-world perturbations may be modeled as shift-scale interventions, in which 
$\smash{f_{\thetab}(\cdot})_j$ and $\smash{\sigma_{\thetab}(\cdot})_j$
of a variable $\smash{x_j}$ are shifted and scaled by  $\smash{\delta,\beta}$, respectively, as
\begin{align}\label{eq:sde-continuous-shift-scale}
\begin{split}
    \hspace*{0pt}
    f_{\thetab,\phib}(\xb)_j = f_{\thetab}(\xb)_j + \delta 
    \hspace*{5pt}\text{and}\hspace*{5pt}
    \sigma_{\thetab,\phib}(\xb)_j = \beta \, \sigma_{\thetab}(\xb)_j \, ,
\end{split}
\end{align}
where $\smash{\phib = \{\delta,\beta\}}$.
Analogous shift interventions have been studied in acyclic and cyclic SCMs \citep{zhang2021matching,rothenhausler2015backshift}.
Most settings assume that interventions are sparse, have known target variables, or few parameters \citep{scholkopf2022causality}.

\vsubsection{Properties} \label{ssec:sdes-causal-models-properties}

\paragraph{Complexity}
Stationary diffusions can be modeled by general functions $f$ and $\sigma$. 
Even with diagonal $\smash{\sigma = \sqrt{2}\Ib}$, they can characterize any observational density $\mu$ via its score function $\smash{f = - \nabla_{\xb} \log \mu}$ (as a Langevin diffusion).
When $\sigma$ is non-diagonal, the driving noise of the equations $\smash{\dxt}$ becomes correlated, which can model confounding.
Thus, the function classes of $f$ and $\sigma$ determine the complexity of the densities modeled by the diffusion, not the Brownian motion $\{\WW_t\}$.
Besides some notable exceptions \citep{immer2022identifiability}, the assumptions of stationary diffusions are less restrictive than those of SCMs, where the exogenous noise defines the distributional family {\em a priori}.

\vparagraph{Stability}
Using diffusions for causal modeling relies on the stationarity, \ie, stability, of the SDEs.
For general $f_{\thetab}$ and $\sigma_{\thetab}$, stability is not guaranteed, particularly when randomly initializing the model parameters~$\thetab$.
For example, in linear systems $\dxt = (\ab + \Bb \xb_t) \dt + \Cb \dWt $, stability requires that the eigenvalues of $\Bb$ have negative real parts \citep{sarkka2019applied}.
Guaranteeing stability under interventions, however, is  
possible in certain cases:
in linear systems, the shift-scale interventions in \eqref{eq:sde-continuous-shift-scale} do not affect stability.
More generally, Theorem~\ref{theorem:consistency} shows that $\mathrm{KDS} = 0$ can guarantee stability and act as a certificate, even for complex model classes.

\vparagraph{Identifiability}
\looseness-1
Causal modeling aims at generalizing to (combinations of) intervention classes when learning a model from a set of observed interventions.
Generalizing to unseen perturbations may not require fully identifying $\thetab$.
For SDEs in particular, a density $\mu$ does not uniquely identify the true parameters $\thetab$ in a model class without unverifiable assumptions:
changing the {\em speed} of a diffusion via $\smash{\dxt = s f(\xb_t)\dt + \sqrt{s} \, \sigma(\xb_t) \dWt}$ for $s > 0$ leaves the stationary density unchanged.
The operator $s \Lcal$ satisfies the same stationarity conditions as~$\Lcal$ \citep{hansen1995back}.
While linear SDEs are identifiable up to speed scaling under specific sparsity conditions \citep{dettling2022identifiability},
it is, to our knowledge, not yet known to what degree multiple interventional densities $\smash{\mu_{\phib}}$ identify stationary SDEs.
As we investigate in Section~\ref{sec:results}, stationary diffusions empirically generalize to unseen interventions, 
hence weaker notions than parametric identifiability may be appropriate.

\vsubsection{Learning Causal Stationary Diffusions from Interventional Data}
\label{ssec:learning-algorithm}

\newlength\mystarlen
\settowidth\mystarlen{$\scriptstyle *$}
\newcommand{\thetabc}{\thetab^*\kern-\mystarlen,\,}

\begin{figure}
    \vspace*{-10pt}
    \centering
    \hspace*{-10pt}\includegraphics[width=1.10\linewidth]{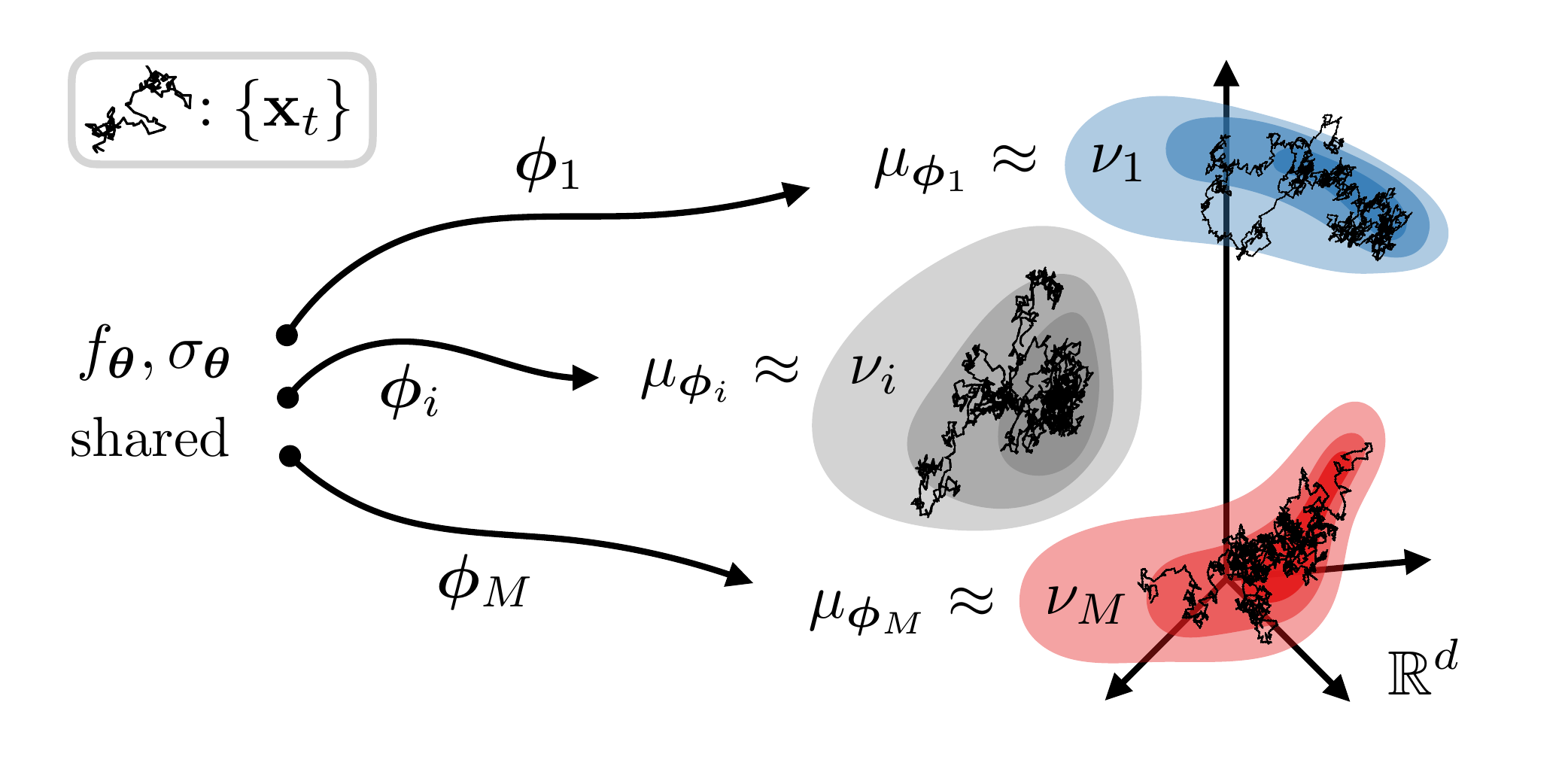}
    \vspace*{-26pt}
    \caption{
    \looseness-1
    {\bf Inference intuition.}
    %
    %
    Our goal is to infer mechanisms     $\smash{f_{\thetab}, \sigma_{\thetab}}$ that explain the observed densities $\nu_{1:M}$. To achieve this, we jointly learn $\thetab$ and interventions $\smash{\phib_i}$ that induce stationary densities $\smash{\mu_{\phib_i}}$ fitting $\nu_i$. 
    \vspace*{-5pt}
    }
    \label{fig:learning}
\end{figure}

Suppose we observe $M$ interventional densities $\nu_i$ of an unknown system $\xb$.
Given corresponding datasets $\smash{D_i \sim \nu_i}$ of i.i.d.\ samples, our goal is to learn a causal model that allows predicting the effects of unseen interventions on the system.
To achieve this, we seek to learn an SDE model $f_{\thetab},\sigma_{\thetab}$, whose stationary densities fit the observed densities $\nu_{i}$ under a considered class of interventions $\phib_i$ (Figure \ref{fig:learning}).
If the model explains all densities $\nu_i$ using one set of causal mechanisms $f_{\thetab}, \sigma_{\thetab}$, adding sparsity and parameter regularization, we may expect it to generalize to unseen interventions.

Using the KDS, the model~$\smash{\thetab}$ and the intervention parameters $\smash{\phib_{1:M}}$ can be learned jointly from $\nu_{1:M}$  with gradient descent. 
We iteratively draw batches of the datasets $\smash{D_{i}}$ and then update $\smash{(\thetab,\phib_i)}$ using the KDS gradients of the intervened SDEs $f_{\thetab,\phib_i}$ and $\sigma_{\thetab,\phib_i}$.
Jointly learning the interventions $\smash{\phib_i}$ alongside the model $\thetab$ is well-posed when we observe multiple $\nu_i$ and the interventions have few degrees of freedom (\eg, sparse targets, few parameters),
since $\thetab$ is shared across all $\smash{\phib_i}$.
To mitigate overfitting, we apply a {\em group lasso} penalty $\smash{\grouplasso(\thetab_j)}$ separately to the parameters $\smash{\thetab_j}$ of each $x_j$, with appropriate groups depending on the model class, to encourage sparse causal dependencies 
\citep{yuan2006model}.
Algorithm~\ref{alg:algo} summarizes the inference method.

When learning both $\smash{f_{\thetab}}$ and $\smash{\sigma_{\thetab}}$, the invariance to speed scaling described in Section~\ref{ssec:sdes-causal-models-properties} can cause an instability close to convergence, as decreasing the speed $s$ via $s\smash{f_{\thetab}}$ and $\sqrt{s}\smash{\sigma_{\thetab}}$
shrinks the KDS.
This can be prevented by fixing (the scale of) subsets of the parameters of $\smash{f_{\thetab}}$ or $\smash{\sigma_{\thetab}}$, for example, the self-regulating dependence of $\smash{f_{\thetab}}(\xb)_j$ on $x_j$.
Empirically, minimizing the KDS was sufficient in combination with sparsity regularization to ensure that the learned SDEs are stable upon convergence.
However, future work could aim at guaranteeing stability directly through properties of the model class \citep{richards2018lyapunov,kolter2019learning}.

\begin{algorithm}[t]
\caption{Learning causal stat.\ diffs.\ via the KDS}
\begin{algorithmic}
\vspace*{2pt}
\STATE{\textbf{Input}:~Interventional datasets 
$\smash{\{D_{1}, \mydots, D_{M}\}}$, \\
\hspace*{34.3pt}kernel~$k$, sparsity penalty $\lambda$, optimizer}
\vspace*{3pt}
\STATE{%
    Initialize model $\smash{\thetab}$ and interventions $\smash{\{\phib_1, \mydots, \phib_M\}}$%
    }
\WHILE{not converged}
    \STATE{Draw environment index $i \sim \mathrm{Unif}(\{1, \mydots, M \})$}
    \STATE{Sample batch of interventional data $D \sim D_i$
    }
    \STATE{%
        Update $\thetab$ and $\phib_i$ with optimizer step 
        $$\displaystyle
        \propto -\nabla_{\thetab,\phib_i} 
        \smash{\Big (
        \hat{\mathrm{KDS}}(\Lcal^{\thetab,\phib_i}, D; k)
        + \lambda 
        \sum_{j=1}^d 
        \grouplasso(\thetab_{j})
        \Big )}
        $$
    }
\ENDWHILE
\STATE{
    \textbf{return} 
    $\smash{\thetab}$ and $\smash{\{\phib_1, \mydots, \phib_M\}}$
}
\vspace*{3pt}
\end{algorithmic}
\label{alg:algo}
\end{algorithm}

\newpage

\vsection{Related Work} \label{sec:related-work}

\paragraph{Causality in dynamical systems}
When observing processes over time, fields like Granger causality \citep{granger1969investigating}, autoregressive modeling \citep{hyvarinen2010estimation}, and system identification \citep{ljung1998system} allow inferring notions of causation.
Recent works leverage continuous optimization for this \citep{pamfil2020dynotears,tank2021neural} or study structure identification in differential equations \citep{bellot2021neural}.
\citet{hansen2014causal} and \citet{peters2022causal} formally study interventions in SDE systems observed over time.
Contrary to these time series settings, we do {\em not} assume observations of a process over time.
Instead, we adopt the novel perspective of using stationary distributions to model and infer causality.
Our approach makes explicit that causal models, including SCMs, are abstractions of processes taking place in time \citep[\eg,][Section 2.3.3]{peters2017elements}---even when causation occurs on scales that either are not or cannot be measured as time series.
\citet{varando2020graphical} also study stationary SDEs in the linear case, but they interpret them as probabilistic graphical models,
not considering causality or interventions.
Orthogonal to our work, \citet{mooij2013from}, \citet{blom2020beyond}, and \citet{bongers2022causal} theoretically investigate how equilibria of differential equations relate to classical SCMs.

\vparagraph{Cyclic graphical modeling}
Several works interpret SCMs in ways that allow cycles 
\citep{richardson1996polynomial,lacerda2008discovering,mooij2011causal,hyttinen2012learning,mooij2013cyclic,rothenhausler2015backshift,sethuraman2023nodags}.
These approaches usually assume additive noise and linearity, sometimes with restrictions on the feedback, and require a unique solution $\xb$ to 
$\xb = f(\xb) + \epsilonb$ given any possible~$\epsilonb$ \citep{bongers2021foundations}.
Our proposal of modeling causality with stationary SDEs shares the intuition of an equilibrium but expands on the insight that cyclicity necessarily introduces a notion of time.
Departing from graphical models ultimately enables us to drop prior model restrictions and consider more general mechanisms and interventions.
As real-world processes evolve in time, some challenge the notion of aggregating causality in graphical models altogether \citep{dawid2010beware,aalen2016can}.
In particular, causal mechanisms that are independent in temporal processes may not be translatable into static 
conditional independencies, and for that matter, graphical models  \citep{tejada2023causal}.

\vparagraph{Statistical inference and kernels}
The idea of producing diffusions that imply certain densities goes back to \citet{wong1964construction},
who linked diffusions with polynomial SDE functions $f$ and $\sigma$ to the Pearson distributions.
In econometrics, the infinitesimal generator and Equation~\eqref{eq:generator-stationary-result} are known tools for fitting diffusion models, but usually with specific parameterizations and test functions (\citealp{hansen1995back,conley1997short,duffie2004estimation}; see \citealp{ait2010operator}, Section 3, for an overview). 
The KDS extends these works by introducing a general-purpose characterization of stationarity that covers an infinite class of test functions in closed form. 
Our techniques establish novel connections between SDEs and RKHSs and build on kernel properties previously used by, for example, kernel mean embeddings \citep{smola2007hilbert}, the MMD \citep{gretton2012kernel}, and the kernelized Stein discrepancy \citep{liu2016kernelized}.

From a statistical perspective, 
the KDS can be viewed as a Stein discrepancy \citep{stein1972bound}.
These discrepancies measure the fit of a distribution $\nu$ to a target $\mu$ by constructing operators $\Scal_\mu$ that produce zero mean functions  $\EE_{\xb\sim\nu}[\Scal_\mu h (\xb)]$ when $\nu = \mu$.
One way of obtaining such an operator $\Scal_\mu$ is through the generator of a stationary Markov process \citep{barbour1988stein}.
The Langevin diffusion $\dxt = \nabla_{\xb} \log \mu(\xb) \dt + \sqrt{2} \dWt $ has the stationary distribution $\mu$,
so one can use \eqref{eq:generator-stationary-result}
to show that
the Stein operator
$(\Scal_\mu h )(\xb) :=  \nabla_{\xb} \log \mu(\xb) h(\xb) + \nabla_\xb h(\xb)$ satisfies $\EE_{\xb\sim\mu}[\smash{\Scal_\mu h} (\xb)] = \mathbf{0}$ for any $h$ \citep{gorham2015measuring}.
By contrast, 
the KDS assigns opposite roles to the operator (now {\em model}, per $\Lcal^{\thetab}$) and the distribution (now {\em target}).
As a result, the KDS enables us to learn general SDEs $f_{\thetab},\sigma_{\thetab}$ inducing a known distribution $\mu$, instead of learning a distribution $\nu$ compatible with a known operator $\smash{\Scal_\mu}$.
This conversely implies that the KDS performs score estimation in the special case of the Langevin diffusion \citep{hyvarinen2005estimation}.

\vsection{Experiments}\label{sec:results}

\newcommand{\wasserstein}{\smash{W_2}} 

\vsubsection{Setup}\label{ssec:results-setup}

The downstream purpose of causal modeling is to predict the effects of interventions in a system.
To evaluate this, we compare the interventional densities predicted by stationary diffusions to those by existing methods.
All methods first learn a causal model from interventional data with known targets and then predict the distributions resulting from unseen interventions by sampling data from the learned models.
The test interventions are out-of-distribution, that is, performed on variables not intervened upon in the training data.

In the following, we summarize the experimental setup.
Appendix \ref{app:experiments} provides additional supplementary details on the data, methods, and metrics for benchmarking.

\vparagraph{Data}
We evaluate the methods on data of sparse cyclic linear systems (SCMs and stationary SDEs) and expression data of sparse gene regulatory networks.
For the latter, we simulate the SERGIO model by \citet{dibaeinia2020sergio}, which requires acyclic dependencies, 
without technical noise.
For each randomly-generated system, we sample observational data and interventional data for \num{10} train and \num{10} test interventions on single, disjoint target variables, each dataset containing \num{1000} observations.
In the linear systems, we perform shift interventions;
in SERGIO, we implement gain-of-function (overexpression)
gene perturbations \citep[\eg,][]{norman2019exploring}.
All datasets are standardized by the mean and variance of the observational data.

\newcommand{\figheightresults}{4.2cm} 

\newcommand{\figvspaceresults}{-11pt} 
\newcommand{\figvspaceresultscaption}{-21pt} 

\newcommand{\fighspaceresultsleft}{-17pt} 
\newcommand{\fighspaceresultsmiddle}{-15pt} 
\newcommand{\fighspaceresultsright}{-20pt} 

\newcommand{\figtextwidthnamed}{0.44\textwidth}
\newcommand{\figtextwidthunnamed}{0.28\textwidth}
\newcommand{\figtextwidthunnamedright}{0.25\textwidth}

\newcommand{\fighspacefirstcaption}{2.7cm} 
\newcommand{\fighspacesecondcaption}{-0.3cm} 
\newcommand{\fighspacethirdcaption}{-0.3cm} 

\newcommand{\fighspacelegend}{-6pt} 


\begin{figure*}[!thb]
    \vspace*{-5pt}
    \centering
    \hspace*{5pt}\includegraphics[width=0.99\linewidth]{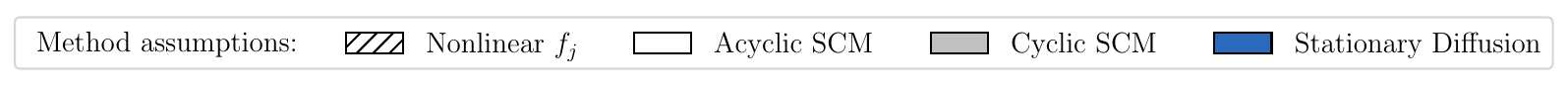}\vspace*{\fighspacelegend}
    
    \hspace{\fighspaceresultsleft}
    \captionsetup[subfigure]{oneside,margin={\fighspacefirstcaption,0cm}}    
    \begin{subfigure}[t]{\figtextwidthnamed}
        \centering
        \subfloat{
            \centering
            \includegraphics[height=\figheightresults]{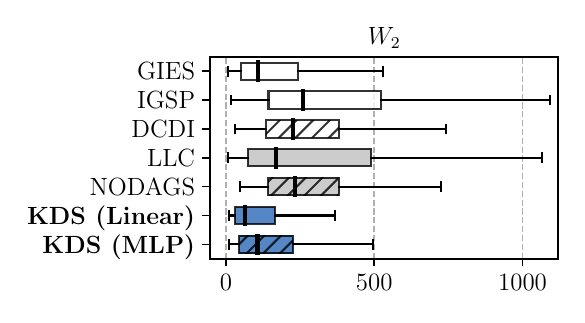}
        }
        \vspace*{\figvspaceresults}

        \subfloat{
            \centering
            \includegraphics[height=\figheightresults]{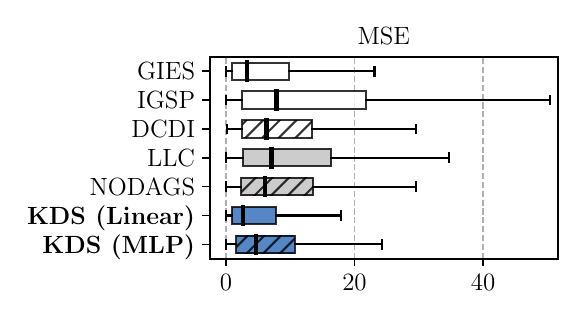}
        }
        \addtocounter{subfigure}{-2}
        \vspace*{\figvspaceresultscaption}
        \caption{{\bf Cyclic Linear SCMs}}\label{fig:results-a}
    \end{subfigure}
    \captionsetup[subfigure]{oneside,margin={\fighspacesecondcaption,0cm}}
    \hfill
    \begin{subfigure}[t]{\figtextwidthunnamed}
        \centering
        \subfloat{
            \centering
            \hspace*{\fighspaceresultsmiddle}\includegraphics[height=\figheightresults]{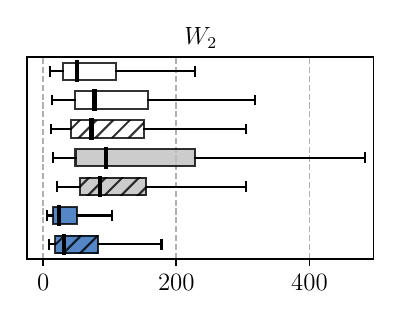}
        }
        \vspace*{\figvspaceresults}

        \subfloat{
            \centering
            \hspace*{\fighspaceresultsmiddle}\includegraphics[height=\figheightresults]{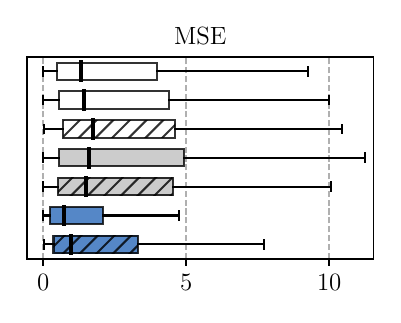}
        }
        \addtocounter{subfigure}{-2}
        \vspace*{\figvspaceresultscaption}
        \caption{{\bf Cyclic Linear SDEs}}\label{fig:results-b}
    \end{subfigure}
    \captionsetup[subfigure]{oneside,margin={\fighspacethirdcaption,0cm}}
    \hfill
    \begin{subfigure}[t]{\figtextwidthunnamedright}
        \centering
        \subfloat{
            \centering
            \hspace*{\fighspaceresultsright}\includegraphics[height=\figheightresults]{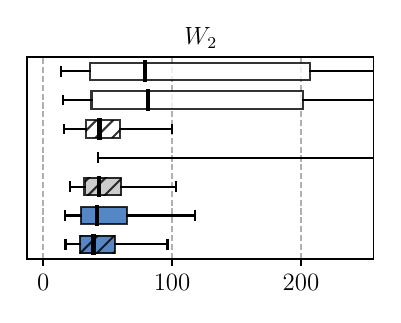}
        }
        \vspace*{\figvspaceresults}

        \subfloat{
            \centering
            \hspace*{\fighspaceresultsright}\includegraphics[height=\figheightresults]{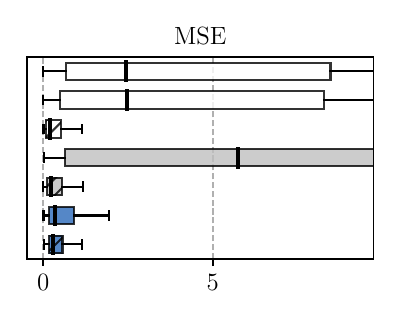}
        }
        \addtocounter{subfigure}{-2}
        \vspace*{\figvspaceresultscaption}
        \caption{{\bf SERGIO}}\label{fig:results-c}
    \end{subfigure}
    \vspace*{-0pt}
    \caption{
    {\bf Benchmarking results} ($\smash{d=20}$ variables, \erdosrenyi causal structure).
    Metrics are computed from \num{10} test interventions on unseen target variables in \num{50} randomly-generated systems.
    Box plots show medians and interquartile ranges (IQR).
    Whiskers extend to the largest value inside \num{1.5} times the IQR length from the boxes.
    Overall, causal stationary diffusions learned via the KDS (Algorithm \ref{alg:algo}, bold-faced) are the most accurate at predicting the effects of interventions on unseen targets, measured in terms of both $\wasserstein$ ($\downarrow$) and MSE ($\downarrow$).
    \vspace*{-0pt}
    }
    \label{fig:results}
\end{figure*}

\vparagraph{Methods}
Using the KDS (Algorithm~\ref{alg:algo}),
we learn stationary diffusions with linear and multi-layer perceptron (MLP) mechanisms $f_{\thetab}(\xb)$,
whose components are defined independently for each $x_j$ as
\begin{align}
    f_{\thetab_j}(\xb)_j 
    &= b^{j} + \wb^{j} \hspace{-1pt}\cdot \xb 
    \tag{Linear} \\
    f_{\thetab_j}(\xb)_j 
    &= b^{j} + \wb^{j} \cdot g(\Ub^{j} \xb + \vb^{j}) - x_j
    \tag{MLP}
\end{align}
where $g$ is the sigmoid nonlinearity.
The diffusion matrix is parameterized as $\smash{\sigma_{\thetab}(\xb) = \diag(\exp(\sigmab))}$.
The interventions $\phib$ that are jointly learned with $\thetab$ to fit the interventional data
are modeled as shifts on the known target variables, as in (\ref{eq:sde-continuous-shift-scale}, left).
Appendix \ref{app:experiments} defines the regularizers $\grouplasso(\thetab_{j})$ of both models.
We use the Gaussian kernel $\smash{k_\gamma}$ to estimate the KDS.
Test-time predictions are sampled from the SDEs via the Euler-Maruyama scheme (Appendix \ref{app:background-euler-maruyama}).

We compare the stationary SDE models to five SCM approaches that use interventional data: score-based GIES \citep{hauser2012characterization}, the constraint-based IGSP algorithm \citep{wang2017permutation}, both based on linear-Gaussian SCMs, and nonlinear DCDI \citep{brouillard2020differentiable}. 
We also benchmark LLC \citep{hyttinen2012learning} and NODAGS \citep{sethuraman2023nodags}, which learn cyclic linear and nonlinear SCMs, respectively. 
The graph discovery methods use a maximum-likelihood linear SCM as the causal model.
The hyperparameters of each method are tuned on validation splits of the interventional datasets later used for learning the evaluated models.

\vparagraph{Metrics}
The test interventions performed to query the learned causal models are shift interventions that match the interventional mean of the target variable in the held-out data.
To allow comparing methods with explicit and implicit densities,
we report the Wasserstein distance $\wasserstein$ between the true interventional data and the data sampled by the learned models under the queried interventions.
We also report the mean squared error (MSE) of the true and predicted empirical means \citep{zhang2022active}.

\vsubsection{Results}\label{ssec:results}

Figures \ref{fig:results-a} and \ref{fig:results-b} present the results for the cyclic linear SCM and stationary SDE systems, respectively.
Both the linear and MLP diffusions learned via the KDS (Algorithm~\ref{alg:algo}) achieve the most accurate interventional density predictions in both the $\wasserstein$ and MSE metrics. 
The acyclic approaches, in particular GIES, show competitive performance,
highlighting a trade-off between model complexity and the entailed inference challenge,
even when the data qualitatively violates acyclicity.
In contrast, the cyclic SCM approaches underperform,
particularly LLC, 
whose model assumptions---apart from standardization of the data---perfectly align with this setting.
The synthetic gene expression data  
assesses all methods under model mismatch.
Figure \ref{fig:results-c}
shows that stationary diffusions, especially the nonlinear MLP diffusion, match the best baselines DCDI and NODAGS, which also model nonlinearity.
This highlights the potential of using stationary SDEs for causal modeling  in complex data-generating processes, even in acyclic settings.
Appendix \ref{app:additional-results} presents additional results for scale-free causal structures, which show similar findings overall, as well as significance tests.
Compared to Figure~\ref{fig:results}, the MLP diffusion achieves worse $\mathrm{MSE}$ in linear systems but remains on par with the baselines.

\vspace{-1pt}
\vsection{Conclusion}\label{sec:conclusion}
\vspace{-2pt}
We propose a new approach for modeling causality and interventional distributions using stationary SDEs.
Similar to real-world processes, stationary diffusions unroll causal dependencies and feedback over time, yet the densities they model remain time-invariant.
To propose a practical algorithm, we derive a tractable kernelized objective for learning stationary SDEs from data.
We believe our results linking diffusions to RKHSs provide new fundamental tools for analyzing and learning diffusions, also beyond the context of causal modeling.
Future directions include formally studying generalization under intervention classes, showing consistency of the KDS for more general models, and learning latent representations governed by stationary diffusions.

\section*{Acknowledgments}
Many thanks to Ya-Ping Hsieh and Mohammad Reza Karimi for the engaging discussions on SDEs in the early stages of this work.
We additionally thank Charlotte Bunne, Pawe\l{} Czy\.{z}, Jonas Rothfuss, and Scott Sussex for their helpful comments on  versions of the manuscript.
This work has also greatly benefited from discussions with Nicolas Emmenegger, Parnian Kassraie, Jonas H{\"u}botter, Mojm{\'i}r Mutn{\'y}, Jonas Rothfuss, Zebang Shen, and Ingo Steinwart, for which we are very thankful.

This research was supported by the European Research Council (ERC) under the European Union's Horizon 2020 research and innovation program grant agreement no.\ 815943 and the Swiss National Science Foundation under NCCR Automation, grant agreement 51NF40 180545.

{
\bibliography{ref.bib}
}

\newpage
\clearpage
\onecolumn
\appendix

\section{Additional Background}\label{app:additional-background}

\subsection{Euler-Maruyama Method}\label{app:background-euler-maruyama}

To approximate the solutions to SDEs, we use the Euler-Maruyama method \citep[][Section 8.2]{sarkka2019applied}.
The Euler-Maruyama approximation of sample paths of the diffusion $\smash{\{\xb_t\}}$ solving $\smash{\dxt = f(\xb_t) \dt + \sigma(\xb_t) \dWt}$ is given by
\begin{align}\label{eq:euler-approximation}
    \xb_{l+1} 
    :=
    \xb_{l} 
    + f(\xb_{l}) \Delta t
    +\sigma(\xb_{l}) \xib_l \sqrt{\Delta t}
\end{align}
for some step size $\Delta t$ and independent vectors $\xib_l \sim \Ncal(\mathbf{0},\Ib)$.
To generate $L$ samples from the stationary density $\mu(\xb)$ of $\smash{\{\xb_t\}}$, we simulate a single sample path and then select every $k$-th state $\xb_{l\cdot k}$ for $\smash{l \in \{1, \mydots, L\}}$ as a sample, where $k$ is a thinning factor as in Markov chain Monte Carlo.
In our experiments, we sample $\xb_0 \sim \Ncal(\mathbf{0}, \Ib)$ and use a step size of $\Delta t = 0.01$, a thinning factor of \num{500}, and \num{100} samples of burn-in, which we selected based on autocorrelation diagnostics of the thinned Markov chains.

\subsection{Sobolev Spaces}\label{app:background-sobolev}

Some of our theoretical results build on the notion of Sobolev spaces.
While not required here, we recommend \citet{adams2003sobolev} for a detailed introduction.
The Sobolev norm  $\smash{\lVert \cdot \rVert_{m,p}}$ of a function $f$ sums the $L_p$ norms of all its partial derivatives up to order $m$ and is defined as
\begin{align*}
 \lVert f \rVert_{m,p} := \Bigg ( \,
 \sum_{\mathbf{n} \in \mathbb{N}^d_0: \lvert \mathbf{n} \rvert \leq m}
 \Big \lVert 
 \frac{\partial^{n_1}}{\partial x_1^{n_1}}
 \dots
 \frac{\partial^{n_d}}{\partial x_d^{n_d}}
 f \Big \rVert_p^p
 \Bigg )^{\nicefrac{1}{p}}
\end{align*}
for $1 \leq p < \infty$. Here, $\smash{\lVert \cdot \rVert_{p}}$ is the $L_p$ norm defined as $\smash{\lVert f \rVert_{p} = \left ( \int_{\RR^d} \lvert f(\xb)\rvert^p \mathrm{d}\xb \right)^{\nicefrac{1}{p}}}$.
The Sobolev space $\smash{W^{m,p}}$ contains all functions $\smash{f: \RR^d \rightarrow \RR}$ such that  $\smash{\lVert f\rVert_{m,p}} < \infty$.
Moreover, the space $\smash{W^{m,p}_c}$ is defined as the closure of $C_c^\infty$ in $\smash{W^{m,p}}$ \citep[][Section 3.2]{adams2003sobolev}.

\subsection{\matern Kernel}\label{app:background-matern}

\newcommand{\varmatern}{r} 

The \matern kernel $\smash{k_{\nu,\gamma}}$ with smoothness and scale parameters $\nu,\sigma > 0$  can be seen as a generalization of the Gaussian kernel that allows controlling the smoothness of the RKHS functions.
We write the \matern kernel $\smash{k_{\nu,\gamma}}(\xb, \xb')$ in terms of the distance $r = \lVert \xb - \xb' \rVert_2$ as 
\begin{align}\label{eq:matern-kernel-general}
    k_{\nu,\gamma}(r) :=
    \tfrac{2^{1-\nu}}{\Gamma(\nu)}
    \left (
    \tfrac{\sqrt{2\nu} \, \varmatern}{\gamma}
    \right )^\nu
    K_\nu
    \left (
    \tfrac{\sqrt{2\nu} \, \varmatern}{\gamma}
    \right ) \, ,
\end{align}
where $\Gamma$ is the gamma function and $\smash{K_\nu}$ is a modified Bessel function of the second kind and order~$\nu$ \citep[][Equation 4.14]{rasmussen2006gaussian}.
Common special cases of $k_{\nu,\gamma}$ have the following explicit forms:
\begin{align*}
    k_{\nu = \nicefrac{1}{2},\gamma}(r) &= 
    \exp \left ( 
    - \tfrac{\varmatern}{\gamma}
    \right ) \\
    k_{\nu = \nicefrac{3}{2},\gamma}(r) &= 
    \left ( 
    1 + \tfrac{\sqrt{3} \varmatern }{\gamma}
    \right ) 
    \exp \left ( 
    - \tfrac{\sqrt{3} \varmatern }{\gamma}
    \right ) \\
    k_{\nu = \nicefrac{5}{2},\gamma}(r) &= 
    \left ( 
    1 + \tfrac{\sqrt{5} \varmatern }{\gamma}
    + \tfrac{5 \varmatern^2 }{3\gamma^2}
    \right ) 
    \exp \left ( 
    - \tfrac{\sqrt{5} \varmatern }{\gamma}
    \right ) 
\end{align*}
The Gaussian kernel $\smash{k_\gamma(r) = \exp(- \varmatern^2 /2\gamma^2)}$ is obtained from $k_{\nu,\gamma}$ as $\nu \rightarrow \infty$.

The following two results will be useful for proving Theorem \ref{theorem:consistency}.
The first statement was originally shown by \citet[][Corollary 10.48]{wendland2004scattered} and follows from \citet[][Equation 4.15]{rasmussen2006gaussian}, linking the \matern RKHS to the Sobolev spaces.
The second result concerns the differentiability of the \matern kernel function:

\begin{lemma}\label{lemma:sobolev-matern-equivalence}
{\normalfont (\citealp{kanagawa2018gaussian}, Example 2.8)}
The RKHS $\Hcal$ of a \matern kernel $\smash{k_{\nu, \gamma}}$ is norm-equivalent to the Sobolev space $\smash{W^{\nu + d/2,2}}$.
Specifically, we have $\smash{h \in \Hcal}$ if and only if $\smash{h \in W^{\nu + d/2,2}}$.
Moreover, there exist constants $c_1,c_2$ such that
$\smash{c_1 \lVert h \rVert_{\nu + d/2,2} \leq \lVert h \rVert_{\Hcal} \leq c_2 \lVert h \rVert_{\nu + d/2,2} }$ for all $h \in \smash{\Hcal}$.
\end{lemma}

\begin{lemma}\label{lemma:differentiability-matern}
{\normalfont (\citealp{stein1999interpolation}, Section 2.7, p.~32)} 
The \matern covariance function $\smash{k_{\nu, \gamma}(r)}$ is $2k$-times differentiable if and only if $\nu > k$.
\end{lemma}

\section{Proofs}\label{app:proofs}

\subsection{Proof of Lemma \ref{lemma:embedding}}

Let $\Bcal_{\mu,\Lcal}: \Hcal \rightarrow \RR$ be the composition of the operators $\EE_{\xb \sim \mu}$ and $\Lcal$  defined as $\Bcal_{\mu,\Lcal} h := \EE_{\xb \sim \mu}[\Lcal h(\xb)]$ for all $h \in \Hcal$.
The functional $\Bcal_{\mu,\Lcal}$ is linear since both the expectation and $\Lcal$ are linear operators.
Moreover, $\Bcal_{\mu,\Lcal}$ is continuous because the functional is bounded on the unit ball of functions in $\Hcal$:
\begin{align*}
    \left \lvert\Bcal_{\mu,\Lcal}h \right \rvert
    &=\left \lvert \EE_{\xb \sim \mu}[\Lcal h(\xb)] \right \rvert &\\
    &\leq\EE_{\xb \sim \mu}[ \left  \lvert \Lcal h(\xb)\right \rvert ]   & \text{Jensen's inequality} \\
    &= \EE_{\xb \sim \mu}\left[ 
    \left  \lvert f(\xb) \cdot \nabla_\xb h(\xb) + \tfrac{1}{2} \tr (\sigma(\xb) \sigma(\xb)^\top \nabla_\xb \nabla_\xb  h(\xb) )
    \right \rvert \right] & \text{by definition} \\
    &\leq \EE_{\xb \sim \mu}\left[ 
    \left  \lvert f(\xb) \cdot \nabla_\xb h(\xb) \rvert + \tfrac{1}{2} \lvert \tr (\sigma(\xb) \sigma(\xb)^\top \nabla_\xb \nabla_\xb  h(\xb) )
    \right \rvert \right] & \text{triangle inequality} \\
    &\leq \EE_{\xb \sim \mu}\left[  
    \lVert f(\xb) \rVert_2 \lVert \nabla_\xb h(\xb)\rVert_2  + \tfrac{1}{2} \lVert \sigma(\xb) \sigma(\xb)^\top \rVert_\mathrm{F} \lVert \nabla_\xb \nabla_\xb  h(\xb)\rVert_\mathrm{\small F} 
    \right] & \text{Cauchy-Schwarz} \\
    &= \EE_{\xb \sim \mu}\bigg[  
    \lVert f(\xb) \rVert_2 \Big ( \sum_{i=1}^d  \big\lvert \tfrac{\partial}{\partial x_{i}} h(\xb) \big\rvert^2 \Big)^{\nicefrac{1}{2}} + \tfrac{1}{2} \lVert \sigma(\xb) \sigma(\xb)^\top \rVert_\mathrm{F} \Big ( \sum_{i=1}^d \sum_{j=1}^d \big\lvert\tfrac{\partial^2}{\partial x_{i} \partial x_{j}} h(\xb) \big\rvert^2\Big)^{\nicefrac{1}{2}}
    \bigg] \hspace*{-80pt}& \\
    &\leq 
    \lVert h \rVert_\Hcal \cdot 
    \EE_{\xb \sim \mu}\bigg [ 
    \lVert f(\xb) \rVert_2 \Big ( \sum_{i=1}^d  \tfrac{\partial}{\partial x_{i,i}} k(\xb, \xb)\Big)^{\nicefrac{1}{2}}  \hspace*{-5pt} + \tfrac{1}{2} \lVert \sigma(\xb) \sigma(\xb)^\top \rVert_\mathrm{F} \Big( \sum_{i=1}^d \sum_{j=1}^d  \tfrac{\partial^2}{\partial x_{i,i} \partial x_{j,j}} k(\xb, \xb)\Big )^{\nicefrac{1}{2}} 
    \bigg ] \, . ~~(*)  \hspace*{-100pt} & 
\end{align*}
The last inequality follows from the fact that the partial derivatives of RKHS functions $h\in\Hcal$ are bounded as
\begin{align*}
    \big\lvert \tfrac{\partial}{\partial x_i} h(\xb)\big\rvert  
    \leq \lVert h \rVert_\Hcal \cdot \big(\tfrac{\partial}{\partial x_{i,i}} k(\xb, \xb)\big)^{\nicefrac{1}{2}} 
    \quad\quad \text{and} \quad\quad
    \big\lvert \tfrac{\partial^2}{\partial x_i\partial x_j} h(\xb)\big\rvert  
    \leq \lVert h \rVert_\Hcal \cdot \big(\tfrac{\partial^2}{\partial x_{i,i}\partial x_{j,j}} k(\xb, \xb)\big)^{\nicefrac{1}{2}}
\end{align*}
\citep[][Corollary 4.36]{steinwart2008support}.
The notation $\smash{\nicefrac{\partial}{\partial x_{i,i}}}$ was defined in Footnote~\ref{footnote:partial}.
By assumption, the squares of $f$, $\sigma$, and the partial derivatives $\smash{\nicefrac{\partial}{\partial x_{i,i}} k(\xb, \xb)}$ and $\smash{\nicefrac{\partial^2}{\partial x_{i,i}\partial x_{j,j}} k(\xb, \xb)}$ are square-integrable with respect to $\mu$, hence the expectation in the above inequality is bounded.
Thus, when applied to the unit ball of $\Hcal$, for which $\smash{\lVert h \rVert_{\Hcal} \leq 1}$, the norm of $\Bcal_{\mu,\Lcal}$ is bounded.
Hence, $\Bcal_{\mu,\Lcal}$ is a continuous linear functional, and by the Riesz representation theorem, there exists a unique representer $g_{\mu,\Lcal} \in \Hcal$ such that
\begin{align} \label{eq:proof-representer-thm}
    \Bcal_{\mu,\Lcal} h = \EE_{\xb \sim \mu}[\Lcal_\xb h(\xb)] = \langle h, g_{\mu,\Lcal} \rangle_\Hcal 
\end{align}
for all $h \in \Hcal$.
We obtain the explicit form for $g_{\mu,\Lcal}$ by substituting $k(\,\cdot\,, \xb') \in \Hcal$ for $h$ in \eqref{eq:proof-representer-thm}, which yields
\begin{align*}
    \EE_{\xb \sim \mu}[\Lcal_\xb k(\xb, \xb')] = \langle k(\,\cdot\,, \xb'), g_{\mu,\Lcal} \rangle_\Hcal  \, ,
\end{align*}
where the notation $\Lcal_\xb$ makes explicit that the operator $\Lcal$ is applied to the argument $\xb$.
By the reproducing property, we have $\langle k(\,\cdot\,, \xb'), g_{\mu,\Lcal} \rangle_\Hcal = g_{\mu,\Lcal}(\xb')$ \citep{scholkopf2002learning}.
Hence, $g_{\mu,\Lcal}(\xb') = \EE_{\xb \sim \mu}[\Lcal_\xb k(\xb, \xb')]$, and we can read off that the representer function is given by $g_{\mu,\Lcal}(\cdot) = \EE_{\xb \sim \mu}[\Lcal_\xb k(\xb, \,\cdot\,)]$. 

\vspace{-12pt}\hfill $\blacksquare$

\subsection{Proof of Theorem \ref{theorem:sup}}

Using the representer function $g_{\mu,\Lcal}$ in Lemma \ref{lemma:embedding}, we can express the supremum over $\Fcal$ as
\begin{align*}
    \sup_{h \in \Fcal}    \EE_{\xb \sim \mu} \big[ \Lcal h (\xb) \big] 
    = \sup_{h \in \Fcal} \langle h, g_{\mu,\Lcal} \rangle_\Hcal
    = \big\langle \tfrac{g_{\mu,\Lcal}}{\lVert g_{\mu,\Lcal} \rVert_\Hcal}, g_{\mu,\Lcal} \big\rangle_\Hcal
    = \lVert g_{\mu,\Lcal} \rVert_\Hcal \, .
\end{align*}
In the above, we used the fact that the norm $\smash{\langle h, g_{\mu,\Lcal} \rangle_\Hcal}$ is maximized over $h \in \Fcal$ by the unit-norm function aligned with $\smash{g_{\mu,\Lcal}}$, that is, by $\smash{g_{\mu,\Lcal} / \lVert g_{\mu,\Lcal} \rVert_\Hcal \in \Fcal}$.
The squared RKHS norm $\smash{\lVert g_{\mu,\Lcal} \rVert_\Hcal^2}$ can be written in terms of the kernel as
\begin{align*}
    \lVert g_{\mu,\Lcal} \rVert_\Hcal^2 
    &= \langle g_{\mu,\Lcal} , g_{\mu,\Lcal} \rangle_\Hcal 
    = \EE_{\xb \sim \mu} \big[ \Lcal_\xb g_{\mu,\Lcal}  (\xb) \big] & \text{Lemma \ref{lemma:embedding}} \\
    &= \EE_{\xb \sim \mu} \big[ \Lcal_\xb \big [ \EE_{\xb' \sim \mu} \big[ \Lcal_{\xb'} k(\xb', \xb) \big] \big]\big] \, .  & \text{Explicit form of $g_{\mu,\Lcal} (\cdot)$} 
\end{align*}
We call this expression the {\em kernel deviation from stationarity} $\smash{\kds}$.

Under additional regularity conditions on $f,\sigma,k,$ and $\mu$, we may interchange the differentials in $\smash{\Lcal_{\xb}}$ with the integral in $\smash{\int \mu(\xb')\Lcal_{\xb'} k(\xb', \xb) \mathrm{d}\xb'}$
(or specifically, their involved limits)
and write
$\smash{\kds} = \smash{\EE_{\xb \sim \mu, \xb' \sim \mu} \big [ \Lcal_{\xb\vphantom{'}}\Lcal_{\xb'\vphantom{'}} k(\xb, \xb') \big ]}$.
For example, by the dominated convergence theorem, one sufficient condition allowing the interchange is when the functions and their first- and second-order partial derivatives are continuous and bounded.
More general conditions are possible.

\vspace{-12pt}\hfill $\blacksquare$

\subsection{Proof of Theorem \ref{theorem:consistency}} \label{app:proof-theorem-consistency}

Let $\Hcal$ be the RKHS of the \matern kernel $k_{\nu,\gamma}$, and let $\Fcal$ be its unit ball.
This proof uses Lemmata \ref{lemma:sobolev-matern-equivalence} and \ref{lemma:differentiability-matern},  two auxiliary results about \matern and Sobolev spaces that are given in Appendix \ref{app:additional-background}.

To begin, we note that $f$, $\sigma$, $\smash{\nicefrac{\partial}{\partial x_{i,i}} k_{\nu,\gamma}(\xb, \xb)}$, and 
$\smash{\nicefrac{\partial^2}{\partial x_{i,i}\partial x_{j,j}} k_{\nu,\gamma}(\xb, \xb)}$ are all square-integrable with respect to $\mu$, because the functions are bounded, and any bounded function is square-integrable with respect to a probability density.
Both functions $f$ and $\sigma$ are bounded by assumption.
Moreover, Lemma~\ref{lemma:differentiability-matern} and $\nu > 2$ imply that the partial derivatives
$\smash{\nicefrac{\partial}{\partial x_{i,i}} k_{\nu,\gamma}(\xb, \xb)}$ and 
$\smash{\nicefrac{\partial^2}{\partial x_{i,i}\partial x_{j,j}} k_{\nu,\gamma}(\xb, \xb)}$ exist and are finite. 
These functions of $\xb$ are bounded, because the \matern kernel function depends only on the distance between its inputs, which is $\smash{\lVert \xb - \xb \rVert_2 = 0}$ for any $\xb$, and thus these partial derivatives are constant with respect to $\xb$.
Given the square-integrability of 
$f$, $\sigma$, $\smash{\nicefrac{\partial}{\partial x_{i,i}} k_{\nu,\gamma}(\xb, \xb)}$, and 
$\smash{\nicefrac{\partial^2}{\partial x_{i,i}\partial x_{j,j}} k_{\nu,\gamma}(\xb, \xb)}$,
all assumptions of Lemma~\ref{lemma:embedding} and Theorem \ref{theorem:sup} are satisfied.

To prove the theorem, we leverage the fact that the smooth functions with compact support $\smash{C_c^\infty}$ form a core for the generator $\Acal$ associated to the SDEs
when $f, \sigma$ are Lipschitz continuous and bounded and the matrix $\sigma(\xb)\sigma(\xb)^\top$ is positive definite for all $\xb \in \RR^d$
\citep[][Theorem 1.6, p.~370]{ethier1986markov}.
%
We can link the core $\smash{C_c^\infty}$ to the \matern RKHS $\Hcal$:

\smallskip

\begin{lemma}\label{lemma:sobolev-matern-dense}
$C^\infty_c$ is a dense subset of $\smash{\Hcal}$ with respect to the Sobolev norm $\lVert \cdot \rVert_{\nu + d/2,2}$. 
\end{lemma}
\looseness-1
\vspace{-5pt}
{\bf Proof of Lemma~\ref{lemma:sobolev-matern-dense}.~~~}%
The space $\smash{W^{m,p}_c}$ is defined as the closure of $\smash{C_c^\infty}$ in the Sobolev space $\smash{W^{m,p}}$ 
(Appendix \ref{app:background-sobolev}).
Therefore, the core $C^\infty_c$ is dense in $\smash{W_c^{m,p}}$ with respect to the Sobolev norm $\lVert \cdot \rVert_{m,p}$.
Moreover, $\smash{W_c^{m,p} = W^{m,p}}$ when both spaces are defined over $\RR^d$ \citep[][Corollary 3.23]{adams2003sobolev}, so $C^\infty_c$ is dense in $\smash{W^{m,p}}$.
From Lemma \ref{lemma:sobolev-matern-equivalence}, we know that $\smash{W^{\nu + d/2,2} = \Hcal}$ for the set of functions.
Hence, $C^\infty_c$ is dense in $\smash{W^{\nu + d/2,2} = \Hcal}$ with respect to the Sobolev norm $\lVert \cdot \rVert_{\nu + d/2,2}$.

\medskip

We now prove both directions of the equivalence in the theorem:
\vspace{-5pt}
\begin{itemize}
    \item[\contour{black}{$\Leftarrow$}~~]

If $\kds = 0$, then $\smash{\sup_{h \in \Fcal} \EE_{\xb \sim \mu} [ \Lcal h (\xb) ] = 0}$ by Theorem~\ref{theorem:sup}.
Since the supremum is nonnegative,
it follows that $\EE_{\xb \sim \mu} [ \Lcal h (\xb) ] = 0$ for all $h \in \Fcal$.
This implies that the equality also holds for $h \in \Hcal$, since the length of the vectors does not affect their orthogonality.
When $\lVert h \rVert_\Hcal > 0$, we can also see this from $\EE_{\xb \sim \mu} [ \Lcal h (\xb) ] = \langle h, g_{\mu,\Lcal} \rangle_\Hcal = \lVert h \rVert_\Hcal \langle h / \lVert h \rVert_\Hcal, g_{\mu,\Lcal} \rangle_\Hcal  = \lVert h \rVert_\Hcal \cdot 0 = 0$
since $ h / \lVert h \rVert_\Hcal \in \Fcal$.

By Lemma \ref{lemma:sobolev-matern-dense}, the core $\smash{C^\infty_c}$ is a subset of $\Hcal$, so we have
$\EE_{\xb \sim \mu} [ \Lcal u (\xb) ] = 0$ for all $u \in \smash{C^\infty_c}$.
If $u \in C^\infty_c$, then $u \in C^2_c$ and thus $\Acal u = \Lcal u$. It follows that $\EE_{\xb \sim \mu} [ \Acal u (\xb) ] = 0$ for all $u$ in the core $\smash{C^\infty_c}$. This implies that $\mu$ is the stationary density \citep[][Chapter 4, Proposition 9.2]{ethier1986markov}.

\item[\contour{black}{$\Rightarrow$}~~]

If $\mu$ is the stationary density, 
we have $\smash{\EE_{\xb \sim \mu} [ \Acal u (\xb) ] = 0}$ for all functions $u$ in the core $\smash{C_c^\infty}$.
Moreover, since $C_c^\infty \subset C^2_c$, it holds that $\EE_{\xb \sim \mu} [ \Lcal u (\xb) ] = 0$. 

Let $h \in \Hcal$. 
By Lemma \ref{lemma:sobolev-matern-dense}, there exists $u \in C^\infty_c$ such that 
$\lVert h - u \rVert_{\nu + d/2,2} < \epsilon$.
By the above, we then have
\begin{align*}
    \lvert \EE_{\xb \sim \mu} [ \Lcal h (\xb) ] \rvert
    &= \lvert \EE_{\xb \sim \mu} [ \Lcal h(\xb) - \Lcal u(\xb) + \Lcal u(\xb) ] \rvert & \text{expanding} \\
    &\leq \lvert \EE_{\xb \sim \mu} [ \Lcal (h-u)(\xb) ] \rvert + \lvert \EE_{\xb \sim \mu} [\Lcal u(\xb) ] \rvert & \text{triangle inequality} \\
    &= \lvert \EE_{\xb \sim \mu} [ \Lcal (h-u)(\xb) ] \rvert \\
    &= \lvert  \langle h-u, g_{\mu,\Lcal} \rangle_\Hcal \rvert & \text{Lemma \ref{lemma:embedding}} \\
    &\leq  \lVert h-u \rVert_\Hcal \, \lVert  g_{\mu,\Lcal} \rVert_\Hcal  & \text{Cauchy-Schwarz} \\
    &\leq  c_2 \lVert h-u \rVert_{{\nu + d/2,2}} \, \lVert  g_{\mu,\Lcal} \rVert_\Hcal  
    & \text{Lemma \ref{lemma:sobolev-matern-equivalence}} \\
    &< c_2 \, \epsilon \, \lVert  g_{\mu,\Lcal} \rVert_\Hcal   
\end{align*}
Thus, $\lvert \EE_{\xb \sim \mu} [ \Lcal h (\xb) ] \rvert$ is bounded by $\epsilon$ times the constants $c_2$ and $\lVert  g_{\mu,\Lcal} \rVert_\Hcal$, which are both independent of the function $h$.
Hence, for all functions $h \in \Hcal$ and any $\epsilon' > 0$, we can choose $\epsilon > 0$ such that $| \EE_{\xb \sim \mu}[\Lcal h(\xb)] | < \epsilon'$.
It follows that $\EE_{\xb \sim \mu} [ \Lcal h (\xb) ] = 0$ for all $\smash{h \in \Hcal}$
and, by Theorem \ref{theorem:sup}, $\kds = 0$.

\hfill $\blacksquare$

\end{itemize}

\section{Additional Details on the Kernel Deviation from Stationarity}\label{app:details-kds}

\subsection{Linear-Time Unbiased Estimator} \label{app:details-kds-linear}

The empirical estimate of $\kds$ in \eqref{eq:kds-sampled} scales quadratically with the size of $\smash{D = \{\xb^{(1)}, \mydots, \xb^{(N)}\}}$.
Similar to \citet{gretton2012kernel},
we can obtain an alternative estimator by subsampling the summands in \eqref{eq:kds-sampled} as
\begin{align}\label{eq:kds-sampled-linear}
    \hat{\mathrm{KDS}}_{\mathrm{linear}}(\Lcal, D; k) := 
    \frac{1}{\lfloor N/2 \rfloor}
    \sum_{m=1}^{\lfloor N/2 \rfloor} 
    \Lcal_{\xb}\Lcal_{\xb'} k(\xb^{(2m-1)}, \xb^{(2m)})  \, .
    \vspace*{-10pt}
\end{align}
The computation of this estimator scales linearly with $N$.
The estimator is unbiased by the linearity of expectations and the fact that samples are i.i.d.\ as $\mu$.
The linear estimator may be advantageous when $N$ is prohibitively large to compute (\ref{eq:kds-sampled}) or when a single evaluation of (the gradient of) $\Lcal^{\thetab}_{\xb}\Lcal^{\thetab}_{\xb'}k(\xb,\xb')$ is expensive, but we do not want to ignore samples from $D$.

\subsection{General Explicit Form and Automatic Differentiation} \label{app:details-kds-explicit}

For general diffusion functions $\sigma$, we cannot use the Laplacian notation of \eqref{eq:generator-explicit-form}, but  $\Lcal^{\thetab}_{\xb}\Lcal^{\thetab}_{\xb'} k(\xb, \xb')$ nevertheless has an explicit form given by
\begin{align}\label{eq:general-kds-form}
\begin{split}
    \Lcal^{\thetab}_{\xb}\Lcal^{\thetab}_{\xb'} k(\xb, \xb') 
    =&~ f_{\thetab}(\xb) \cdot \nabla_{\xb} \nabla_{\xb'} k(\xb, \xb')  \cdot f_{\thetab}(\xb')\\
    &+ \tfrac{1}{2} f_{\thetab}(\xb) \cdot \nabla_{\xb} \tr \big( \sigma_{\thetab}(\xb) \sigma_{\thetab}(\xb)^\top \, \nabla_{\xb'}\nabla_{\xb'} k(\xb, \xb') \big)\\
    &+ \tfrac{1}{2} f_{\thetab}(\xb') \cdot \nabla_{\xb'} \tr \big( \sigma_{\thetab}(\xb') \sigma_{\thetab}(\xb')^\top \, \nabla_{\xb}\nabla_{\xb} k(\xb, \xb') \big)\\
    &+ \tfrac{1}{4} \tr \Big( \sigma_{\thetab}(\xb) \sigma_{\thetab}(\xb)^\top \,\, \nabla_{\xb}\nabla_{\xb} \tr \big( \sigma_{\thetab}(\xb') \sigma_{\thetab}(\xb')^\top \nabla_{\xb'}\nabla_{\xb'} k(\xb, \xb') \big) \Big) \, .
\end{split}
\end{align}
Precomputing the kernel terms is still possible to some degree, but fewer steps of the computation can be cached in the trace terms.
In this case, it may thus be simpler implementation-wise to compute the parameter gradients $\nabla_{\thetab} \Lcal^{\thetab}_{\xb}\Lcal^{\thetab}_{\xb'} k(\xb, \xb')$
and
$\nabla_{\phib} \Lcal^{\thetab, {\phib}}_{\xb}\Lcal^{\thetab,{\phib}}_{\xb'} k(\xb, \xb')$
with automatic differentiation at each update step in Algorithm~\ref{alg:algo}.

\newcommand{\codevspaceup}{-27pt}
\newcommand{\algoillustrationrelwidth}{0.65}

\makeatletter
\newcommand*\mysize{%
  \@setfontsize\mysize{7.0}{8.3}%
}
\makeatother

\definecolor{codegreen}{rgb}{0,0.6,0}
\definecolor{codegray}{rgb}{0.5,0.5,0.5}
\definecolor{codepurple}{rgb}{0.58,0,0.82}
\definecolor{backcolour}{rgb}{0.95,0.95,0.92}
\definecolor{lightgray}{rgb}{0.7,0.7,0.7}

\definecolor{codeblue}{rgb}{0.13,0.49,0.83} 


\lstdefinestyle{mystyle}{
    commentstyle=\color{codegreen},
    keywordstyle=\color{codeblue},
    numberstyle=\tiny\color{codegray},
    stringstyle=\color{codepurple},
    basicstyle=\ttfamily\mysize,
    breakatwhitespace=false,         
    breaklines=true,                 
    captionpos=b,                    
    keepspaces=true,                 
    numbers=left,                    
    numbersep=5pt,                  
    showspaces=false,                
    showstringspaces=false,
}

\lstset{style=mystyle}

\makeatletter
\newcommand\currentStyle@lstparam{}
\lst@AddToHook{Output}{\global\let\currentStyle@lstparam\lst@thestyle}
\lst@AddToHook{OutputOther}{\global\let\currentStyle@lstparam\lst@thestyle}
\makeatother

\makeatletter
\newcommand{\highlightcode}[2]{\currentStyle@lstparam \textcolor{#1}{#2}}
\makeatother

\newcommand{\codeempty}{\vphantom{X}}
\newcommand{\emphh}{\highlightcode{codegray}{h}}
\newcommand{\empharg}{\highlightcode{codegray}{arg}}

\makeatother

\begin{figure}[!t]
    \centering
    \begin{minipage}[t]{\algoillustrationrelwidth\textwidth}
    \begin{center}
    \vspace*{\codevspaceup}
    \vspace*{15pt} 
    \hspace*{15pt}\begin{minipage}[t]{\textwidth}
    \input{algo/code_long}
    \end{minipage}
    \vspace*{-10pt}
    \end{center}
    \vspace*{5pt}
    \end{minipage}
    \captionof{figure}{\looseness-1 
     Computing the 
     generator 
     gradient
     $\smash{\nabla_{\thetab} \Lcal^{\thetab}_{\xb}\Lcal^{\thetab}_{\xb'}k(\xb,\xb')}$
     via
     two calls of the operator
     $\smash{\Lcal^{\thetab}}$ (here: {\small\texttt{L}}).
     }
     \label{fig:autodiff-code}
\end{figure}

To make this concrete,
Figure \ref{fig:autodiff-code}
provides
Python pseudocode using \mytexttt{autograd} and
\mytexttt{JAX}-style syntax \citep{jax2018github}
that illustrates how to compute $\smash{\nabla_{\thetab} \Lcal^{\thetab}_{\xb}\Lcal^{\thetab}_{\xb'} k(\xb, \xb')}$ in only a few lines.
Following \eqref{eq:differential-operator},
the operator $\Lcal^{\thetab}$ can be conveniently defined as a higher-order function that accepts a function $h$ as input and returns the function $\Lcal^{\thetab}h$.
After applying $\Lcal^{\thetab}$ once to each argument of $k(\xb, \xb')$, the function  \mytexttt{kds} in Figure \ref{fig:autodiff-code} exactly computes \eqref{eq:general-kds-form}, and we can directly compute its gradient with respect to $\thetab$ in the last line using automatic differentiation.

\section{Experimental Setup} \label{app:experiments}

\subsection{Data}\label{app:experiments-data}

\subsubsection{Causal Structures}\label{app:sparsity-structure}
For benchmarking, we simulate data from randomly-generated sparse linear systems and sparse gene regulatory network models with $d=20$ variables.
Following prior work (\eg, \citealp{zheng2018dags}), 
we sample random causal structures $\Gb \in \{0,1\}^{d \times d}$ with the number of causal dependencies per variable following a polynomial or power-law distribution, corresponding to \erdosrenyi and scale-free graphs, respectively. 
\erdosrenyi graphs are sampled by drawing links independently with a fixed probability (when acyclic, restricted to an upper-triangular matrix).
Scale-free graphs are generated by  preferential attachment, where links $j$ to the previous $\smash{j-1}$ nodes are sampled with probability proportional to its degree and then randomly directed (when acyclic, always directed ingoing to~$j$).
For both structure distributions, we fix the expected degree of the variables to \num{3}.

\subsubsection{Cyclic Linear Systems}\label{app:data-linear-systems}

\vparagraph{Models}
Given some $\Gb \in \{0,1\}^{d \times d}$,
we generate random instances of the two cyclic linear models
\begin{align}
    \xb &= \Wb\xb + \bb + \diag(\sigmab) \epsilonb ~~\text{with}~ \epsilonb \sim \Ncal(\mathbf{0}, \Ib) \tag{Cyclic Linear SCM}\\[5pt]
    \dxt &= (\Wb\xb_t + \bb) \dt + \diag(\sigmab) \dWt \tag{Cyclic Linear SDE}
\end{align}
where $\smash{\Wb \in \RR^{d \times d}, \bb \in \RR^{d}}$, and $\smash{\sigmab \in \RR^{d}_{\scriptscriptstyle > 0}}$,
and $\Wb$ is sparse according to $\smash{\Gb}$.
Sampling cyclic systems requires more caution than in the acyclic case, since both generative processes must be stable.
For SCMs, the maximum of the real-parts of the eigenvalues $\rho(\Wb)$ must be less than \num{1}, for SDEs less than \num{0}.
For an insightful evaluation, we additionally want $\Wb$ to be asymmetric and not approximately diagonal, \ie, have significant causal dependencies between the variables.

To generate such systems, we first sample $\smash{\Gb \sim p(\Gb)}$, $\smash{\Wb \sim p(\Wb)}$, $\smash{\bb \sim p(\bb)}$, $\smash{\sigmab \sim p(\sigmab)}$. Then, we
multiply $\Wb$ times $\Gb$ elementwise along their offdiagonal elements and finally subtract $\smash{\rho(\Wb) + \epsilon}$ from the diagonal of $\Wb$, which ensures that $\smash{\rho(\Wb) \leq -\epsilon}$.
This protocol empirically induced stronger variable correlations in the stationary distributions than the procedure by \citet{varando2020graphical}.
They perform a more vacuous diagonal shift based on the Gershgorin circle theorem, often resulting in large dominating diagonals. 
For our experiments, we use 
$\smash{p(w_{ij}) = \mathrm{Unif}(-3, -1)\cup(1, 3)}$ 
and $\epsilon = 0.5$ for the matrices 
and ${p(b_j) = \mathrm{Unif}(-3, 3)}$ 
and ${p(\log \sigma_j) = \mathrm{Unif}(-1, 1)}$ for the biases and scales, respectively, both for the SCMs and SDEs.
To sample the SCM data, we draw $\epsilonb \sim \Ncal(\mathbf{0}, \Ib)$ and then compute $\smash{\xb = (\Ib - \Wb)^{-1}(\bb + \diag(\sigmab)\epsilonb)}$ \citep{hyttinen2012learning}.
To sample from the stationary density of the SDEs,
we use the Euler-Maruyama scheme (Appendix \ref{app:background-euler-maruyama}).

\vparagraph{Interventions}
\looseness-1
Given the fully-specified linear model,
we sample an observational dataset and interventional data for single-variable shift interventions on all variables, each with \num{1000} observations, as the interventions for the benchmark.
In both SCMs and SDEs, the shift intervention is implemented by adding a scalar $\delta$ to the bias $\smash{b_j}$ of the target variable $j$.
In our experiments, we sample $\delta \sim p(\delta)$ with $\smash{p(\delta) = \mathrm{Unif}(-15, -5)\cup(5, 15)}$ 
independently for each intervention.

\subsubsection{Gene Regulatory Networks}

\vparagraph{Model}
Given some acyclic $\Gb \in \{0,1\}^{d \times d}$,
we use the SERGIO model by \citet{dibaeinia2020sergio} and their corresponding implementation (GNU General Public License v3.0) 
to sample synthetic gene expression data.
The gene expressions are simulated by a stationary dynamical system over a sparse, acyclic regulatory network encoded by $\Gb$.
To simplify the experimental setup, we use the clean gene expressions without technical measurement noise as the observations.

SERGIO models the mRNA concentration of the genes using the chemical Langevin equation, a nonlinear geometric Brownian motion model driven by two independent Wiener processes for each gene.
The expression $x_j$ of gene $j$ is primarily defined through its production rate $p_j$, which depends nonlinearly on the expression levels $\xb$ of the other genes through the signed interaction parameters $\Kb$
and the regulatory network $\Gb$.
Following \citet{dibaeinia2020sergio}, we use a Hill nonlinearity coefficient of \num{2} and sample the parameters $k_{ij}$ as well as \num{10} master regulator rates $b_{jc}$, which model cell type heterogeneity, from $k_{ij} \sim \mathrm{Unif}(-5, -1)\cup(1, 5)$ and $b_{jc} \sim \mathrm{Unif}(1, 4)$, respectively.
Finally, we use an expression decay rate of $\lambda = 0.5$ and noise scale of $q = 0.5$, which deviates from the values $0.8$ and $1.0$, respectively, used by  \citet{dibaeinia2020sergio} when simulating $d \geq 100$ genes.
Under their settings, the data of smaller networks does not contain sufficient signal for the any of the benchmarked methods to learn a nontrivial model of the system.

\vparagraph{Interventions}
Given the fully-specified gene regulation model,
we sample an observational (wild-type) dataset and interventional data for single-variable gain-of-function (overexpression) interventions \citep[\eg,][]{norman2019exploring} on all genes, each with \num{1000} measured cell observations, as the interventions for the benchmark.
We evaluate overexpression rather than knockdown perturbations, because the former are qualitatively more similar to the test-time shift interventions used to query the models learned by the methods.
The gain-of-function interventions are implemented by multiplying the production rate $p_j$ of the target gene $j$ by a randomly-sampled factor $r_j \sim \mathrm{Unif}(2, 10)$.
The half-response levels for the Hill nonlinearities are kept at the values estimated during the wild-type simulation, so that the intervention effects propagate downstream.

\subsection{Metrics}

We focus on comparing the true and predicted interventional distributions of unseen interventions in a system.
This evaluation setting mimics applications, but benchmarking different algorithms requires some care.
In general, there is a mismatch between the perturbation implemented by an intervention in the ground-truth system and the {\em query} perturbation performed in a learned causal model%
---not only because the true model perturbation is unknown, but also because true and learned models may be from different model classes.

\vparagraph{Test-time interventions}
To compare different models at test-time,
we query each learned model by a shift intervention on the target variable that induces the same target variable mean as the true, held-out interventional data \citep{rothenhausler2015backshift,zhang2021matching}.
After performing the intervention, our metrics compare the predicted and true interventional joint distributions.
We perform shift interventions, because they have the same definition in both SCMs \eqref{eq:scm} and stationary diffusions \eqref{eq:sde-continuous-shift-scale}. For both, we add a scalar $\delta$ to the mechanism $f_j(\xb)$ of the target variable~$j$ (see Appendix \ref{app:data-linear-systems}).
To make the query well-defined, we assume knowledge of the true interventional mean of the target variable.

For acyclic SCMs, the query shift $\smash{\delta}$ is given by the difference between the empirical observational mean of the learned SCM and the target interventional mean.
However, cyclic SCMs and stationary diffusions may model feedback on the target variable, where the above does not hold.
For these models,
we find the query shifts $\delta$ by an exponential search around $\delta = 0$ for a range estimate $(\delta_\mathrm{lo}, \delta_\mathrm{hi})$.
For each shift, we simulate data from the intervened model and compare the predicted empirical mean to the target mean of the intervened variable. 
Given an estimated range $(\delta_\mathrm{lo}, \delta_\mathrm{hi})$, we run a grid search for the optimal value 
$\delta \in 
\{\delta_\mathrm{lo},
\delta_\mathrm{lo} + \nicefrac{1}{10}(\delta_\mathrm{hi} - \delta_\mathrm{lo}), 
\mydots,
\delta_\mathrm{lo} + \nicefrac{9}{10}(\delta_\mathrm{hi} - \delta_\mathrm{lo}),
\delta_\mathrm{hi}\}$.
Ultimately, we select the shift $\delta$ achieving the minimum distance to the target interventional mean.

\vparagraph{Metrics}
Both metrics we report are computed based on samples from the interventional distributions, which enables a nonparametric comparison across the different models.
For each test intervention, we simulate \num{1000} samples $\tilde{\xb}^{(n)} \in \tilde{D}$
from the interventional distribution of the predicted model and compare them with the true interventional dataset of \num{1000} samples 
$\xb^{(n)} \in D$.

To evaluate the overall fit of the predicted distribution, we compute the Wasserstein distance $\wasserstein$ to the ground-truth interventional data.
To make $\wasserstein$ efficiently computable, we report the $\wasserstein$ distance with small entropic regularization,
which interpolates between $\wasserstein$ and the MMD \citep{genevay2018learning}.
The entropy-regularized $\wasserstein$ distance between the empirical measures of the datasets 
$D$ and $\tilde{D}$
with $\lvert D \rvert = M$ and $\lvert \tilde{D} \rvert = N$
is defined as
%
%
\begin{align*}
    \wasserstein(D, \tilde{D})  := 
    \left (
    \,\,
    \min_{\substack{
    \vspace*{0pt}\\
    \textstyle \Pb \in U
    }}
    \,
    \sum_{m=1}^M
    \sum_{n=1}^N
    \,
    p_{mn} \lVert 
    \xb^{(m)} - \tilde{\xb}^{(n)}
    \rVert_2^2
    - \epsilon H[\Pb]
    \,\,
    \right )^{\nicefrac{1}{2}} \, ,
\end{align*}
where $H$ is the entropy defined as 
$\smash{H[\Pb] := - \sum_{nm} p_{mn}(\log p_{nm} - 1)}$,
and $U$ is the set of transport matrices
$\smash{U = \{ \Pb \in \RR^{M \times N}_{\scriptscriptstyle \geq 0} :~  \Pb \mathbf{1}_N = \nicefrac{1}{M} \, \mathbf{1}_M ~\text{and}~ \Pb^\top \mathbf{1}_M = \nicefrac{1}{N} \, \mathbf{1}_N  \}}$ with $\mathbf{1}_N$ being a vector of $N$ ones
\citep{peyre2019computational}.
For evaluation, the $\wasserstein$ metric is more robust than the MMD, because it does not depend on the sensitive choice of a kernel bandwidth \citep{gretton2012kernel}.
For $\epsilon > 0$, $\wasserstein(D, \tilde{D})$ can be efficiently computed using the Sinkhorn algorithm. We use the \mytexttt{ott-jax} package (Apache 2.0 Licence) and $\epsilon = 0.1$ \citep{cuturi2022optimal}.

In addition to the overall fit, we separately assess the accuracy of the interventional means alone.
Following \citet{zhang2022active}, we report the mean squared error of the  predicted empirical means of the $d$ variables given by
\begin{align*}
    \mathrm{MSE}(D, \tilde{D})
    :=
    \frac{1}{d}
    \sum_{j=1}^d 
    (m_j - \tilde{m}_j
    )^2 \, ,
\end{align*}
where 
$\smash{\mb := \tfrac{1}{M} \sum_{m=1}^M \xb^{(m)}}$ 
and 
$\smash{\tilde{\mb} := \tfrac{1}{N} \sum_{n=1}^N \tilde{\xb}^{(n)}}$
are the empirical means of the datasets.

\subsection{Hyperparameter Tuning}\label{app:hyperparameter-tuning}

In the experiments, we benchmark the methods on different generative processes (Appendix~\ref{app:experiments-data}).
To calibrate the important hyperparameters of the methods, we perform cross-validation prior to the final evaluation that benchmarks the methods.
All methods are tuned separately for each data-generating process, that is, for cyclic linear SCMs, cyclic linear SDEs, and the gene expression data, for \erdosrenyi and scale-free sparsity structures.

The experiments for all generative processes are repeated for \num{50} randomly-sampled systems.
For each system, we generate an observational and \num{10} interventional datasets for learning a model as well as \num{10} interventional datasets for evaluation, with all interventions performed on different target variables.
To tune the hyperparameters of the methods, 
we split the \num{10} observed interventions into \num{9} training and \num{1} validation dataset.
The methods then infer a causal model based on the \num{9} training interventional and the observational dataset, 
and we compute the $\wasserstein$ metric for the unseen validation intervention. 
For each method, we select the hyperparameter configuration achieving the lowest median $\wasserstein$ metric on \num{20} randomly-selected tasks.

\subsection{Stationary Diffusions}\label{app:experiments-ours}

\vparagraph{Models}
We evaluate linear and nonlinear stationary diffusion models. 
Both classes of SDE systems model $d$ independent drift and diffusion mechanisms $\smash{f_j}$ and $\smash{\sigma_j}$
that are defined by separate parameters $\smash{\thetab_j}$.
For both models, the corresponding group lasso regularizers $R(\thetab_j)$ penalize the dependence on the other variables.
The models and regularizers are defined as
\begin{align*}
&\begin{aligned}
    f_{\thetab_j}(\xb)_j 
    &= b^{j} + \wb^{j} \hspace{-1pt}\cdot \xb \\
    f_{\thetab_j}(\xb)_j 
    &= b^{j} + \wb^{j} \cdot g(\Ub^{j} \xb + \vb^{j}) - x_j    
\end{aligned}
&&
\begin{aligned}
    \grouplasso(\thetab_j) 
    &= \textstyle \sum_{i\neq j}^d \lvert w^j_i \rvert \\
    \grouplasso(\thetab_j) 
    &= \textstyle \sum_{i\neq j}^d \lVert \ub^j_i \rVert_2  
\end{aligned}
\end{align*}
where $g(z) := \exp(z) / (\exp(z) + 1)$ the sigmoid nonlinearity, applied elementwise.
The diffusion term $\sigma$ is modeled as a constant diagonal matrix $\sigma(\xb) = \diag(\exp(\log \sigmab))$, with $\log \sigmab \in \RR^d$.
The parameters $\log \sigmab$ are learned in log-space to enable gradient-based optimization.
To remove the speed scaling invariance, we fix $\smash{w^j_j = -1}$ in the linear and $\smash{\ub_j^j = \mathbf{0}}$ in the MLP model (see Section \ref{ssec:learning-algorithm}).
In the experiments, the MLP model uses a hidden size of $h = 8$ for the matrices $\Ub^{j} \in \RR^{h \times d}$ and vectors $\vb^j, \wb^j \in \RR^h$.

\vparagraph{Interventions during training}
For both SDE models, we model the drift mechanisms $\smash{f_{\thetab, \phib}(\xb)_j}$ of the variables targeted by interventions as shift interventions as defined in (\ref{eq:sde-continuous-shift-scale}, left) with parameters $\smash{\phib_j = \{\delta_j\}}$.
As described in Algorithm \ref{alg:algo}, we learn the parameters $\smash{\phib_j}$ jointly with $\thetab$, since they are unknown.
For the purpose of the experiments, we limit the learned interventions to shifts in order to allow a direct comparison with SCMs.
However, we found learning more complex intervention parameterizations like, for example, full shift-scale interventions as in \eqref{eq:sde-continuous-shift-scale}, generally straightforward. 
More expressive interventions shift some of the burden of explaining the distribution shift from $\thetab$ to $\phib_j$, which can help inferring robust parameters $\thetab$ under model mismatch.

\vparagraph{Optimization}
For all experiments, we run \num{20000} update steps on the KDS as described in Section \ref{ssec:learning-algorithm}
using the Adam optimizer with learning rate \num{0.001}.
We compute the empirical KDS \eqref{eq:kds-sampled} using the Gaussian kernel $\smash{k_\gamma}$ and a batch size of $\lvert D \rvert = 512$. 
For initialization of the parameters $\thetab$, we use a zero-mean Gaussian for the linear model and LeCun-Uniform initialization for the nonlinear model, respectively, both with scale $0.001$.
We initialize the intervention shifts $\phib_j = \{ \delta_j \}$ by warm-starting them at the difference in means of the target variable in the interventional and the observational datasets.
Overall, the important hyperparameters are the kernel bandwidth $\gamma$ and the group lasso regularization strength $\lambda$, so we tune these for each experimental setting using the protocol described in Appendix \ref{app:hyperparameter-tuning}.

\vparagraph{Diagnostics}
The following intuitions may be helpful when deploying our inference approach.
If the stationary density induced by the learned SDEs overfits or collapses to a small part of the data, the kernel bandwidth may be too small.
Our bandwidth range is suitable for standardized datasets of $d=20$ variables but should likely be expanded in different settings.
If the learned SDEs are unstable upon convergence or diverge during simulations---despite a decreasing or near-zero KDS loss---then the speed scaling invariance may not be adequately fixed (see above and Section \ref{ssec:learning-algorithm}).
In this context, we find that the fit and performance of the models empirically improves when fixing the self-regulating parameters of $f_j$ on $x_j$, rather than, \eg, the noise scales $\sigma_j$.
Without any sparsity regularization, 
Algorithm \ref{alg:algo} may converge to models at the edge of stability, \eg, to linear models with maximum real parts of the eigenvalues being near zero and only just negative.
Sparsity regularization can mitigate such instability and related issues.

\subsection{Baselines}\label{app:baselines}

\begin{table}[t]
    \centering
    \caption{{\bf Hyperparameter tuning for the experiments in Section \ref{sec:results}.} The hyperparameters of all methods are selected using the protocol described in Appendix \ref{app:hyperparameter-tuning}.}\label{tab:hyperparameters}
    \vspace{5pt}
\begin{adjustbox}{max width=\linewidth}
\begin{threeparttable}
\begin{tabular}{lll}
\toprule
\textbf{Method} & \textbf{Hyperparameter} & \textbf{Search range} \\
\midrule
IGSP 
& significance level & $\alpha_\mathrm{IGSP} \in \{ 0.001, 0.003, 0.01, 0.03, 0.1 \}$ \\
\midrule
DCDI 
& sparsity regularization & $\lambda_{\mathrm{DCDI}} \in \{0.001, 0.01, 0.1, 1, 10 \}$ \\
& number of MLP layers & $ m_{\mathrm{DCDI}} \in \{1, 2 \} $ \\
\midrule
NODAGS 
& sparsity regularization & $\lambda_{\mathrm{NODAGS}} \in \{0.0001, 0.001, 0.01, 0.1\}$ \\
& spectral norm terms & $n_{\mathrm{NODAGS}} \in \{5, 10, 15\}$ \\
& learning rate & $\eta_{\,\mathrm{NODAGS}} \in \{ 0.001, 0.01, 0.1 \}$ \\
& hidden units & $m_{\mathrm{NODAGS}} \in \{1, 2, 3\}$ \\
\midrule
LLC 
&  sparsity regularization & $\lambda_{\mathrm{LLC}} \in \{ 0.001, 0.01, 0.1, 1, 10, 100\}$ \\
\midrule
KDS 
& sparsity regularization & $\lambda \in \{0.001, 0.003, 0.01, 0.03, 0.1 \}$ \\
& kernel bandwidth & $\gamma \in \{ 3, 5, 7\}$ \\
\bottomrule
\end{tabular}
\end{threeparttable}
\end{adjustbox}
\end{table}

\textbf{GIES} \citep{hauser2012characterization} assumes a linear-Gaussian SCM to infer a graph equivalence class, from which we randomly sample a causal graph. 
To perform the greedy search, we run the original \mytexttt{R} implementation of the authors using the Causal Discovery Toolbox (MIT Licence)\footnote{\href{https://github.com/FenTechSolutions/CausalDiscoveryToolbox}{\mytexttt{https://github.com/FenTechSolutions/CausalDiscoveryToolbox}}}.
Given the DAG estimate, we use a linear-Gaussian SCM with maximum likelihood parameter and variance estimates as the learned model.
These estimates have simple closed-forms that account for interventional data \citep{hauser2012characterization}. 
At test time, the shift interventions are implemented in the learned linear SCM and the data sampled as described in Appendix \ref{app:data-linear-systems}.

\smallskip
\textbf{IGSP} \citep{wang2017permutation} uses a Gaussian partial correlation test. We use the same closed-form maximum likelihood parameter and variances estimates as for GIES to construct the final causal model.
For IGSP, we run the implementation provided as part of the CausalDAG package (3-Clause BSD License)\footnote{\href{https://github.com/uhlerlab/causaldag}{\mytexttt{https://github.com/uhlerlab/causaldag}}}.
Using the protocol described in Appendix \ref{app:hyperparameter-tuning}, we tune the significance level $\alpha_\mathrm{IGSP}$ of the conditional independence test for each experimental setting individually by searching over 
a range of $\alpha_\mathrm{IGSP}$ values (see Table \ref{tab:hyperparameters}).
As for GIES, the shift interventions are implemented in the estimated linear SCM as described in Appendix \ref{app:data-linear-systems}.

\smallskip
\textbf{DCDI} \citep{brouillard2020differentiable} learns a nonlinear, Gaussian SCM parameterized by neural networks jointly with the noise variance.
For comparison with the nonlinear stationary diffusion model, we use the same hidden size of \num{8} for the neural networks. 
To run DCDI, we use the Python implementations provided by the authors (MIT License).
We tune the regularization strength $\lambda_{\mathrm{DCDI}}$ and the number of layers $m_{\mathrm{DCDI}}$ and leave the remaining hyperparameters at the suggestions by the authors (see Table \ref{tab:hyperparameters}).
When learning from imperfect interventions, DCDI estimates a separate model for each interventional environment. 
For evaluation, we use the model learned for the observational dataset and implement the shift interventions by adding the bias $\delta$ to the mean of the Gaussian modeling the target variable, analogous to the linear SCMs and SDEs described in Appendix \ref{app:data-linear-systems}.

\smallskip
\textbf{NODAGS} \citep{sethuraman2023nodags} infers a nonlinear cyclic SCM using residual normalizing flows and also estimates the noise variances.
As suggested by the authors, we jointly tune 
the regularization parameter $\lambda_{\mathrm{NODAGS}}$,
the number of terms for computing the spectral norm $n_{\mathrm{NODAGS}}$,
the learning rate $\eta_{\,\mathrm{NODAGS}}$,
and the number of hidden units $m_{\mathrm{NODAGS}}$
(see Table \ref{tab:hyperparameters}).
We set the remaining hyperparameters to the recommendations by the authors and use the implementation published alongside the original paper (Apache 2.0 Licence). 
At test time, the shift interventions are implemented in the model by using $\Ub = \Ib$ and otherwise as described in the paper, analogous to the linear cyclic SCMs described in Appendix \ref{app:data-linear-systems}  (see also \citealp{hyttinen2012learning}).

\smallskip
\textbf{LLC} \citep{hyttinen2012learning} learns a linear cyclic SCM and estimates the noise variances.
For the basic implementation of the LLC algorithm, we use the code provided by the NODAGS repository. 
However, we extend their implementation by the $\ell_1$ sparsity regularizer described in Section 6.2 of the original paper by \citet{hyttinen2012learning}, solving the minimization problem with BFGS.
We treat the weight $\lambda_{\mathrm{LLC}}$ of this regularizer as a hyperparameter that is tuned via a grid search (see Table \ref{tab:hyperparameters}).
At evaluation time, the shift interventions in the learned cyclic linear SCM are performed as for GIES and IGSP.

\subsection{Compute Infrastructure}\label{app:compute-infrastructure}

The development and experiments of this work were carried out on an internal cluster.
In each experiment, all methods ran for up to \num{1} hour of wall time on up to \num{4} CPUs and \num{16} GB of RAM, adjusted individually according to the compute requirements of each method.
We implement our approach with
\mytexttt{JAX} \citep{jax2018github} and thus additionally provide \num{1} GPU, which allows for significant speed-ups during development and the experiments.
Overall, running our inference method takes approximately one hour given the above resources, 
both for the linear and nonlinear model,
and including the final search for test-time intervention shifts.

\section{Additional Results} \label{app:additional-results}

\begin{figure*}
    \centering
    \hspace*{5pt}\includegraphics[width=0.99\linewidth]{img/legend.pdf}\vspace*{\fighspacelegend}

    \hspace{\fighspaceresultsleft}
    \captionsetup[subfigure]{oneside,margin={\fighspacefirstcaption,0cm}}
    \begin{subfigure}[t]{\figtextwidthnamed}
        \centering
        \subfloat{
            \centering
            \includegraphics[height=\figheightresults]{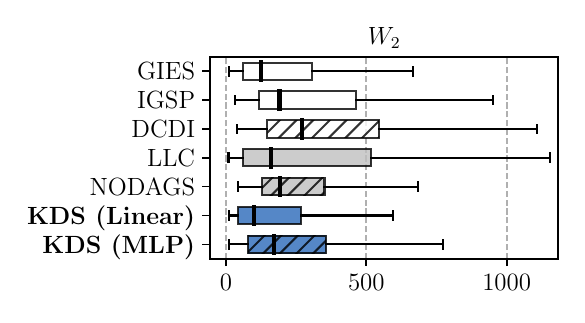}
        }
        \vspace*{\figvspaceresults}

        \subfloat{
            \centering
            \includegraphics[height=\figheightresults]{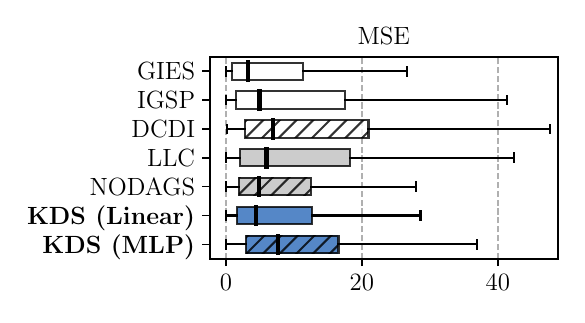}
        }
        \addtocounter{subfigure}{-2}
        \vspace*{\figvspaceresultscaption}
        \caption{{\bf Cyclic Linear SCMs}}\label{fig:results-sf-a}
    \end{subfigure}
    \captionsetup[subfigure]{oneside,margin={\fighspacesecondcaption,0cm}}
    \hfill
    \begin{subfigure}[t]{\figtextwidthunnamed}
        \centering
        \subfloat{
            \centering
            \hspace*{\fighspaceresultsmiddle}\includegraphics[height=\figheightresults]{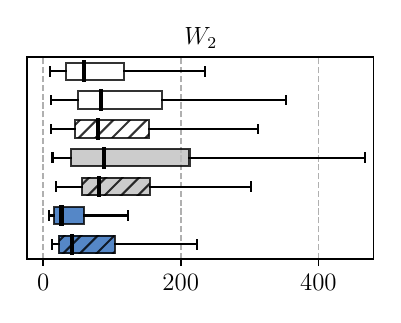}
        }
        \vspace*{\figvspaceresults}

        \subfloat{
            \centering
            \hspace*{\fighspaceresultsmiddle}\includegraphics[height=\figheightresults]{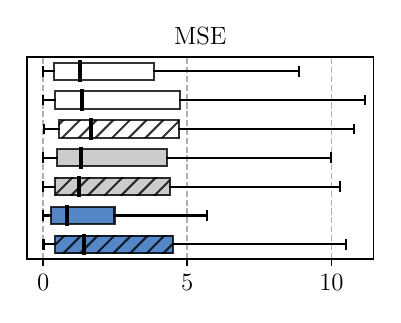}
        }
        \addtocounter{subfigure}{-2}
        \vspace*{\figvspaceresultscaption}
        \caption{{\bf Cyclic Linear SDEs}}\label{fig:results-sf-b}
    \end{subfigure}
    \captionsetup[subfigure]{oneside,margin={\fighspacethirdcaption,0cm}}
    \hfill
    \begin{subfigure}[t]{\figtextwidthunnamedright}
        \centering
        \subfloat{
            \centering
            \hspace*{\fighspaceresultsright}\includegraphics[height=\figheightresults]{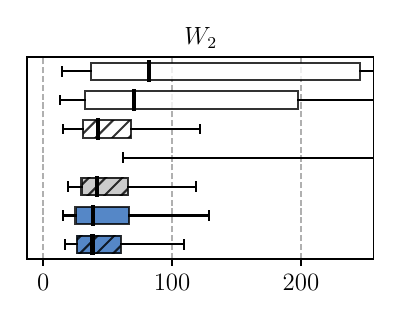}
        }
        \vspace*{\figvspaceresults}

        \subfloat{
            \centering
            \hspace*{\fighspaceresultsright}\includegraphics[height=\figheightresults]{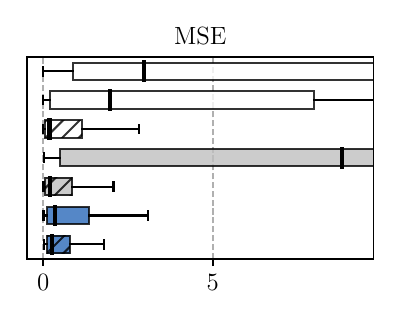}
        }
        \addtocounter{subfigure}{-2}
        \vspace*{\figvspaceresultscaption}
        \caption{{\bf SERGIO}}\label{fig:results-sf-c}
    \end{subfigure}
    \vspace*{2pt}
    \caption{
    {\bf Benchmarking results} ($\smash{d=20}$ variables, scale-free causal structure).
    Metrics are computed from \num{10} test interventions on unseen target variables in \num{50} randomly-generated systems.
    Box plots show medians and interquartile ranges (IQR).
    Whiskers extend to the largest value inside \num{1.5} times the IQR length from the boxes.
    }
    \label{fig:results-sf}
\end{figure*}

Figure \ref{fig:results-sf} presents supplementary experimental results. The setup is the same as in Figure \ref{fig:results}, except that the causal dependency structure of the ground-truth systems is scale-free rather than \erdosrenyi (see Appendix \ref{app:sparsity-structure}).
Similar to Figure \ref{fig:results}, the stationary diffusion models, in particular those with linear mechanisms $f_{\thetab}$, are the most accurate at predicting the effects of the unseen interventions overall. The MLP variants achieve slightly worse results compared to Figure \ref{fig:results} but are still competitive with the nonlinear SCM baselines. 
Here, stationary diffusions may further improve if the sparsity regularizer described in Appendix \ref{app:experiments-ours} groups out- rather than ingoing dependency parameters, inducing power law out- rather than in-degree degree structure.

Since some of the reported metrics exhibit high variance, we additionally ran Wilcoxon signed-rank tests (one-sided, significance level $\alpha = 0.05$).
The results are shown in Table \ref{tab:tests}.
We find that our approach significantly outperforms the baselines across most metrics and data settings.

\newpage

\begin{table}
\vspace*{5pt}
\caption{
{\bf Significance tests supporting the experimental results. }
The table shows ($*$/$*$) if the alternative hypothesis---our approach outperforming the baselines on average, \ie, the metric distribution being in our favor---was accepted for either metric ($W_2$/$\mathrm{MSE}$) using a Wilcoxon signed-rank tests (one-sided, significance level $\alpha = 0.05$).
}\label{tab:tests}
\vspace{7pt}
\centering
\begin{threeparttable}
\begin{adjustbox}{max width=0.4\linewidth}
\begin{tabular}{l ll ll ll}
& \multicolumn{3}{c}{\textbf{KDS (Linear)}}
 & \multicolumn{3}{c}{\textbf{KDS (MLP)}}
\\
\cmidrule{1-7}
Figure
 & \multicolumn{1}{c}{\ref{fig:results-a}}
 & \multicolumn{1}{c}{\ref{fig:results-b}}
 & \multicolumn{1}{c}{\ref{fig:results-c}}
 & \multicolumn{1}{c}{\ref{fig:results-a}}
 & \multicolumn{1}{c}{\ref{fig:results-b}}
 & \multicolumn{1}{c}{\ref{fig:results-c}}
\\
\cmidrule{1-7}
GIES   & $*$/$*$ & $*$/$*$ & $*$/$*$ & $*$/    & $*$/$*$ & $*$/$*$ \\
IGSP   & $*$/$*$ & $*$/$*$ & $*$/$*$ & $*$/$*$ & $*$/$*$ & $*$/$*$ \\
DCDI   & $*$/$*$ & $*$/$*$ & $*$/    & $*$/$*$ & $*$/$*$ & $*$/  \\
LLC    & $*$/$*$ & $*$/$*$ & $*$/$*$ & $*$/$*$ & $*$/$*$ & $*$/$*$ \\
NODAGS & $*$/$*$ & $*$/$*$ & $*$/    & $*$/$*$ & $*$/$*$ & $*$/  \\
\cmidrule{1-7}
\end{tabular}
\end{adjustbox}
\hspace*{1.2cm}
\begin{adjustbox}{max width=0.4\linewidth}
\begin{tabular}{l ll ll ll}
& \multicolumn{3}{c}{\textbf{KDS (Linear)}}
 & \multicolumn{3}{c}{\textbf{KDS (MLP)}}
\\
\cmidrule{1-7}
Figure
 & \multicolumn{1}{c}{\ref{fig:results-sf-a}}
 & \multicolumn{1}{c}{\ref{fig:results-sf-b}}
 & \multicolumn{1}{c}{\ref{fig:results-sf-c}}
 & \multicolumn{1}{c}{\ref{fig:results-sf-a}}
 & \multicolumn{1}{c}{\ref{fig:results-sf-b}}
 & \multicolumn{1}{c}{\ref{fig:results-sf-c}}
\\
\cmidrule{1-7}
GIES   & $*$/    & $*$/$*$ & $*$/$*$ & \hphantom{$*$}/  & $*$/   & $*$/$*$ \\
IGSP   & $*$/$*$ & $*$/$*$ & $*$/$*$ & $*$/             & $*$/   & $*$/$*$ \\
DCDI   & $*$/$*$ & $*$/$*$ & $*$/    & $*$/$*$          & $*$/   & $*$/    \\
LLC    & $*$/$*$ & $*$/$*$ & $*$/$*$ & $*$/             & $*$/   & $*$/$*$ \\
NODAGS & $*$/$*$ & $*$/$*$ & $*$/    & $*$/             & $*$/   & $*$/    \\
\cmidrule{1-7}
\end{tabular}
\end{adjustbox}
\end{threeparttable}
\vspace*{0pt}
\end{table}

\end{document}